# Solution space path planning: A real-time human-centered path planning algorithm for en-route air traffic control

Yiyuan Zou *, Wenying Lyu, Clark Borst

*Control and Simulation, Faculty of Aerospace Engineering, Delft University of Technology, Delft, the Netherlands*

ABSTRACT

As technology advances, various algorithms have been proposed for air traffic management, yet their operational adoption in tactical control remains limited. This gap motivates a human-centered design emphasizing algorithmic interpretability, controller-relevant operational constraints, and real-time computation. Inspired by the interpretability and flexibility of solution space displays, as well as by the decision logic controllers naturally apply when enforcing operational constraints, this study extends the solution space concept to path planning and develops a fast conflict-free path-planning algorithm for en-route Air Traffic Control (ATC), termed Solution Space Path Planning (SSPP). The algorithm integrates three intent-based conflict detection methods—distance-based, time-interval-based, and zone-based—within the solution space framework to identify conflict-free paths in computationally efficient ways. SSPP is developed using both vertex-based and edge-based search nodes, resulting in two variants—SSPPV and SSPPE, respectively. Empirical results show that SSPPV paired with zone-based conflict detection achieves the lowest runtime, computing paths in 3.20 ms on average in a Dutch sector using a 2.5 nmi grid. SSPPV remains approximately 27.49 times faster than SSPPE on average across the conflict detection methods while offering competitive effectiveness, making it suitable for time-critical operations and uninterrupted 'what-if' probing in real time. An extension to SSPPV and SSPPE further examines the trade-off between delay minimization and separation requirements, demonstrating the flexibility of SSPP in revising optimization objectives. This study not only introduces a novel path-planning algorithm but also shows how such algorithms can be designed to align with human use and operational requirements, supporting their integration into future ATC systems.

## 1. Introduction

In response to the increasing demands for efficiency and safety, various technological initiatives have been developed to facilitate automation within the Air Traffic Management (ATM) community. As outlined in the European ATM master plan [1], future ATM is expected to involve increasingly close collaboration between human and machine agents. At higher levels of automation, system performance depends not only on the ability of algorithms to generate timely and reliable solutions, but also on whether these solutions are operationally meaningful and can be understood and supervised by human operators [2–4].

Despite significant advances, the operational adoption of automated decision-support systems remains challenging. This is because advanced algorithms may lack transparency and interpretability, fail to fully capture operational contexts and constraints, or be misaligned with the way Air Traffic Controllers (ATCos) perform their tasks [5]. These considerations motivate a human-centered approach in which algorithm design explicitly accounts for both operational requirements and how feasible actions and constraints are represented to human operators [2,6].

A relevant foundation is the solution space concept, which originated from Ecological Interface Design [7–9] and represents feasible actions as geometrically interpretable "go/no-go" regions bounded by physical and procedural constraints [10]. Previous studies have demonstrated the effectiveness of this concept in supporting real-time human decision-making in Air Traffic Control (ATC) [11] and many other domains (see [12] for an overview). Recently, this concept has been applied in real-world ATC operations in the Dutch airspace [13]. However, existing solution space displays generally support relatively simple decision problems, such as conflict resolution with a single intermediate waypoint, which limits the range of solutions that can be explored.

Motivated by these observations, this paper extends the solution space concept from interface design to the design of a conflict-free path-planning algorithm from a human-centered perspective. In this study, "human-centered" refers to a design approach: the algorithm is struc-tured around an explicit (visual) representation of feasible actions and operational constraints relevant to human ATCos, while accounting for human factors requirements in terms of response times due to computational and rendering delays on the order of milliseconds to support uninterrupted real-time interaction [14]. The term "interpretability" refers to *intrinsic interpretability*, as commonly defined in interpretable machine learning and explainable AI [15,16], where the internal decision process of an algorithm can be directly inspected and traced. SSPP follows a similar design philosophy: its planning process is represented through an explicit search tree, while the solution space reveals the constraints that determine which actions are feasible. Note that this notion is distinct from *perceived interpretability* or actual human understanding [17], which require empirical evaluation with human operators.

* Corresponding author.
*E-mail addresses:* y.zou@tudelft.nl (Y. Zou), w.lyu-1@tudelft.nl (W. Lyu), c.borst@tudelft.nl (C. Borst).

By combining the interpretable solution space representation with search-based path planning, the proposed algorithm can efficiently explore a rich set of conflict-free actions while incorporating operational constraints such as separation assurance, maneuverability limits, and route practicality. The resulting design therefore aims to balance intrinsic interpretability, computational efficiency, operational compliance, and solution quality.

To limit the research scope, we focus on en-route ATC, where the primary task is to guide aircraft from entry to exit points within sectors while avoiding conflicts among them. The resulting problem is formulated as conflict-free path planning, and the proposed algorithm is termed Solution Space Path Planning (SSPP). Compared with existing studies, the main contributions of this paper can be summarized as follows:

1) *Interpretability and operational compliance*: We propose SSPP as an intrinsically interpretable path-planning framework, in which the planning process can be visually traced through an explicit search tree and the solution space represents the operational constraints governing feasible actions. SSPP also overcomes the limitations of traditional A*-based algorithms by generating straighter paths with fewer turns, which are particularly suitable and practical for aircraft navigation. Compared to previous relevant work in aviation [18–20], SSPP is capable of finding conflict-free paths with multiple intermediate waypoints while maintaining the paths as straight as possible. Moreover, the solution space concept inherently provides flexibility in incorporating various constraints, enabling algorithm designers to tailor SSPP to meet different operational requirements.
2) *Computational efficiency*: We develop several advanced techniques to accelerate solution space generation and exploration, paving the way for ATCo 'what-if' probing in real time. For example, to check whether a certain location is unreachable due to *static* obstacles, we implement the recursive shadowcasting algorithm [21,22] for computing the *field of view* instead of relying on a traditional line-of-sight approach [23,24]. To account for turning angle limitations of aircraft, the original shadowcasting algorithm is adapted by incorporating the angle of view constraint (e.g., $[\theta - 60^\circ, \theta + 60^\circ]$ where $\theta$ is the current heading of aircraft). For *dynamic* obstacles, we develop three *intent-based* conflict detection methods for solution space generation: distance-based, time-interval-based, and zone-based. Integrating zone-based conflict detection as used in ecological ATC displays with shadowcasting enables the construction of a field of view that accounts for dynamic obstacles (not just static obstacles), offering a novel and efficient approach to conflict-free path planning in dynamic environments. Empirical evaluation shows that the zone-based approach is generally faster than both the distance-based and time-interval-based approaches, particularly when the grid size is small.
3) *Solution quality*: We implement A* search [25], rather than greedy search, and propose vertex-based and edge-based search nodes for path planning, leading to two variants of SSPP—SSPPV and SSPPE. SSPPE can achieve a higher success rate than SSPPV in finding conflict-free solutions by exploring a larger search space, although this comes at the cost of increased computational time. Since the basic objective of SSPP is to minimize flight time (or delay), we further explore a potential extension that incorporates aircraft separation into the objective to improve solution robustness.

The paper is structured as follows: Section 2 reviews related work, with an emphasis on the application of the solution space concept and existing path-planning algorithms. Section 3 outlines the necessary preliminaries, clarifying the path-planning problem in en-route ATC. Section 4 describes how to generate solution spaces using three conflict detection methods, which lays the foundation for the proposed solution space path planning introduced in Section 5. A case study is presented in Section 6, followed by a discussion of the results in Section 7. Section 8 concludes the article and outlines directions for future research.

## 2. Related work

### 2.1. Solution space concept

In area control, ATCos are responsible for ensuring the safe and efficient movement of aircraft between sectors. They continuously monitor and manage traffic in real time to prevent conflicts, enforce airspace procedural constraints, and maintain optimal flow. However, increasing traffic volumes and the shift toward advanced operational concepts—such as stepping away from fixed geographical boundaries and moving to time-based trajectory management—have significantly increased the complexity of tactical decision-making. This increasing complexity, coupled with the need for more effective automation support, has driven the development of the solution space concept in decision support tools [26], which have recently reached operational implementation. For example, in 2024, the Maastricht Upper Area Control Centre (MUAC) developed and implemented the Lateral Obstacle & Resolution Display (LORD) for conflict resolution support [13].

The solution space refers to the set of feasible actions that ensure safety while meeting operational constraints. Instead of directly presenting solutions, the portrayed envelope highlights the constraints that define these feasible actions, allowing controllers to 'see' the available options and make their own decisions within the boundaries set by these constraints. Building on this concept, various tools have been developed for supporting humans in performing ATC tasks (e.g., conflict detection and resolution) [27], such as the Highly Interactive Problem Solver (HIPS), the Solution Space Diagram (SSD), and the Travel Space Representation (TSR). The SSD, based on velocity obstacles [28–30], portrays heading and speed constraints around the velocity vector of the selected aircraft and within the minimum and maximum speed envelope [31–33]—a concept that also underlies MUAC's LORD system [13]. HIPS shows geographical locations of conflict zones along the planned trajectory, distinguishing between safe and unsafe paths by trajectory extrapolation [11,34]. TSR portrays conflict-free regions for placing intermediate waypoints within an envelope of allowable time delays, particularly for rerouting tasks that empower the controller to manipulate flight trajectories while adhering to spatial-temporal constraints [35–37].

As shown in Fig. 1, when a controlled aircraft is approaching a target waypoint but is at risk of conflict with an intruder aircraft, SSD, HIPS, and TSR each provide different visualizations of the situation. Both SSD and HIPS highlight conflict zones in red—SSD with respect to the velocity vector, and HIPS along the absolute trackline. For both the SSD and HIPS, the ATCo can resolve the conflict by changing the aircraft's heading and/or speed. TSR similarly highlights unsafe regions in red, but these regions now indicate the set of unsafe intermediate waypoints. The key distinction lies in how TSR represents conflict-free options: it considers a series of potential rerouting segments that temporarily deviate from the original flight path to bypass the conflict before returning to the planned route. This enables controllers to perceive safe intermedi-ate waypoint choices, supporting both conflict resolution and rerouting tasks. The TSR is primarily designed for scenarios that involve rerouting via a *single* intermediate waypoint, a simplification that enhances usability and interpretability. ATCos are free to select any point within the safe zone, with the main trade-off being additional track miles and potential time delays.

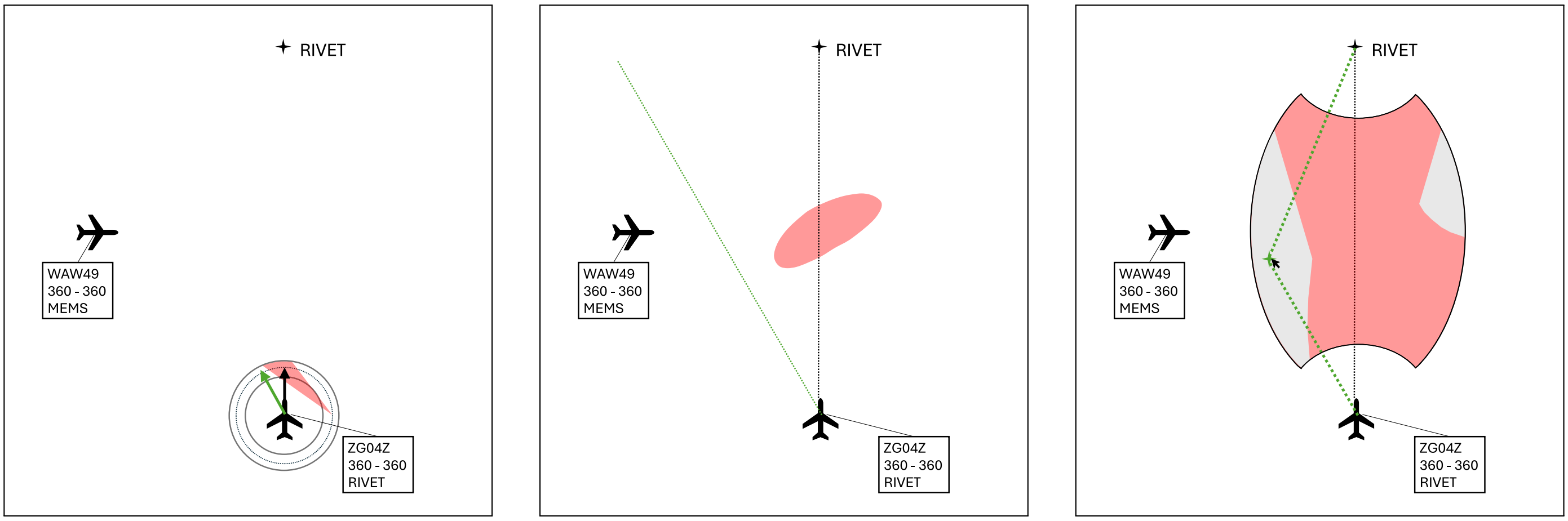

(a) SSD. (b) HIPS. (c) TSR.

**Fig. 1.** Comparison of SSD, HIPS, and TSR representations.

Empirical studies have demonstrated that these tools are effective in supporting ATCos, largely due to their high interpretability and intuitive visual representations [10,26,33,38]. However, they do not necessarily lead to improved operational efficiency and often fail to yield optimal solutions. Consequently, another line of research has emerged, focusing on the development of advanced automation, such as conflict-free path-planning algorithms, for supporting optimal control and decision-making. This increasing level of automation however often comes at the cost of reduced interpretability and transparency, making such algorithms more difficult for human operators to understand and trust [17,39]. This limitation hinders their operational applicability, particularly in safety-critical domains. The trade-off between optimization and acceptance remains a central challenge in the design of decision-support systems [5].

Therefore, building on the foundation of interpretable solution space interfaces—which have found their way into operational use—this paper introduces Solution Space Path Planning (SSPP). The proposed approach aims to achieve low-delay routing with fast computation while preserving the intrinsically interpretable structure of the solution space. Whether these design properties translate into perceived interpretability or improved controller understanding is not evaluated through a user study in this article. However, prior work has indicated that intrinsically interpretable path-planning algorithms and their associated visualizations can improve human understanding [17].

### *2.2. Path planning algorithms*

A* [25] is one of the most well-known path-planning algorithms, which can find optimal paths within predefined graphs efficiently, like regular grids, visibility graphs and navigation meshes [40,41]. It has been applied for fixed-wing aircraft path planning [42]. However, the paths found by traditional A* are highly affected by the graph shape. In regular square grids, the turning angles of A* paths are limited to 45-degree increments, such as 45°, 90°, and 135°, due to its 8 adjacent neighbors. This may lead to an excessive number of turning points and overly large turning angles in paths, making the traditional A* unsuitable for crewed aircraft. Some research attempted post-hoc smoothing to straighten the final paths [43], but this approach may not always be effective when multiple constraints are present.

To address this issue, any-angle path planning was proposed [23, 44–46], which ignores the edges of predefined graphs and allows any-angle turns at the vertices. Instead of performing post-hoc smoothing, any-angle path planning searches for smooth paths directly during the planning process. Some techniques have been developed for any-angle path planning. For example, rather than 8 neighbors, one can use extended $2^k$ ($k > 3$) neighbors for node expansion [46]. In this way, the A* search can skip the adjacent grid cells to find straighter paths. Alternatively, one can examine whether a newly expanded node can be reached by the parent of the current node, like Theta* [23]. Any-angle path planning has been applied for fixed-wing aircraft as well [47].

Nonetheless, the aforementioned algorithms are only applicable to static obstacles. They cannot avoid conflicts between aircraft. For conflict-free path planning, the simplest approach may be to utilize TSR. Since TSR evaluates all potential rerouting points, such as each cell in a grid, one can directly choose the point with the lowest cost (e.g., minimum delay) [19]. However, this approach is limited to finding paths with a single turning point, making it less effective in complex scenarios. To relax this restriction, A* can be adapted to consider dynamic obstacles as well. A common approach is to divide the time dimension into multiple equal-length time slots, creating a space-time grid [18,20,48]. Dynamic obstacles can thus be mapped onto this space-time grid and converted into static obstacle cells. Then, A* can be used to find conflict-free paths within this space-time grid. However, this method introduces a new dimension to the search space, which aggravates the *curse of dimensionality* problem [49] (i.e., the algorithm speed may drop significantly as the dimension increases). To address this issue, Safe Interval Path Planning (SIPP) [50] was proposed. Rather than using a space-time grid, SIPP computes safe intervals for each cell, indicating periods during which the cell is unoccupied by obstacles and thus safe for path planning. It avoids discretizing the time dimension and can generate a *compact* space-time search space. SIPP has been applied for aircraft taxiing operations [51].

However, SIPP faces similar problems as A*. The turning angles of SIPP paths are also limited to 45-degree increments. Therefore, combining both SIPP and any-angle path planning, Any-Angle SIPP (AA-SIPP) was developed [52]. Similar to Theta*, AA-SIPP checks whether the parent of the current node can reach a newly expanded node at a lower cost. This smoothing takes dynamic obstacles into account as well. However, also like Theta*, AA-SIPP cannot guarantee finding optimal paths. Time-Optimal AA-SIPP (TO-AA-SIPP) [24] was thus proposed. Instead of merely straightening the path locally around the current node, TO-AA-SIPP examines the entire search space and globally constructs paths that are as straight as possible. This operation could be highly time-consuming [24,53].

To speed up TO-AA-SIPP, Zeta*-SIPP [53,54] was developed. In Zeta*-SIPP, a new type of forward expansion tailored for optimal any-angle path planning was introduced: *elliptical forward expansion*. Elliptical forward expansion does not proceed in a grid-by-grid or $2^k$-grid manner; instead, it expands nodes following an elliptical pattern. By leveraging the properties of ellipses, it can be proven that this elliptical expansion is sufficient for optimal any-angle path planning. Another technique introduced in Zeta*-SIPP is the *field of view*. It is a more efficient approach than the line of sight for detecting conflicts with static obstacles and building visibility connections between nodes.

The SIPP-based algorithms mentioned above are all single-agent methods, which can serve as low-level planners for multi-agent path planning [52,55,56]. The first-come-first-served strategy [19], commonly used in aviation, is well-suited for prioritized planning [51,52]. This means that aircraft already within the sector are considered dynamic obstacles for newly incoming aircraft. In this study, we also implement this strategy. Optimal multi-agent pathfinding methods, such as Conflict-Based Search (CBS) [57] and Any-Angle Continuous CBS (AA-CCBS) [56], are beyond the scope of this article.

In addition to the search-based algorithms discussed above, there is a rapidly growing interest in machine learning-based and intelligent optimization methods for path planning [58]. Recent studies have applied reinforcement learning and hybrid intelligent approaches to autonomous marine vehicle path planning in unknown, obstacle-rich, and cooperative environments [59–61], demonstrating their potential for complex path-planning problems.

However, in aviation, such methods still face significant challenges regarding their trustworthiness and certification [62]. Their inherent "black-box" nature may hinder acceptance and trust from ATCos. Some efforts have been made to mitigate this issue through explainable AI and increased transparency [39]. For example, Herman [63] employed an explainable reinforcement learning technique [64] that decomposes the overall reward into multiple meaningful components, thereby revealing the underlying rationale behind each maneuvering behavior. Papadopoulos et al. [65] introduced operational transparency, which aims to reveal relevant operational information to help ATCos maintain situation awareness and intervene in a timely manner. Zou and Borst [3,6] proposed a unified transparency taxonomy to integrate operational and engineering transparency, although they focused on search-based algorithms rather than learning-based approaches.

Despite these ongoing efforts, there is still a long way to go before machine learning-based methods can be applied in operational ATC environments. Therefore, this study centers on search-based path planning, whose explicit and deterministic state-transition and search-tree structures allow the planning process to be directly inspected and traced. These properties provide a suitable foundation for developing an intrinsically interpretable path-planning algorithm.

The core of conflict-free path planning is conflict detection. In aviation, the Closest Point of Approach (CPA) is the most commonly used metric, whereas safe intervals are mainly employed in SIPP-based path planning. Therefore, we integrate distance-based (CPA) and time-interval-based (SIPP) conflict detection methods into SSPP for comparison. Inspired by HIPS, we also propose a novel conflict detection method for path planning based on conflict zones (zone-based conflict detection). This method inherits the explicit geometric representation of conflict constraints in HIPS and is fully compatible with the field-of-view scans. Moreover, we develop two versions of SSPP and introduce a new objective function for the trade-off between delay and separation.

Compared to existing path-planning algorithms, the proposed SSPP is similar to the naive TO-AA-SIPP [24]. At each step, SSPP generates a solution space to indicate which nodes are reachable from the current node. However, unlike TO-AA-SIPP, the solution space in SSPP incorporates real-world constraints, and the field of view is leveraged to facilitate its generation. Another closely related study was conducted by Chen et al. [19], which similarly proposes a path-planning algorithm for en-route airspace operations and employs the solution space concept to support controller decision-making. In their approach, *distance-based conflict detection* is used to construct the solution space, while the search space is restricted to a *single* intermediate rerouting point. Under this setting, SSPP-D (where D denotes distance-based conflict detection) with only one rerouting point follows essentially the same search logic.

## 3. Preliminaries

In this section, we clarify the path-planning problem in en-route ATC. To simplify the problem, we make the following assumptions:

1) *Consider only 2D instead of 3D*. As the airspace is divided into multiple flight levels, it is reasonable to focus only on 2D within each flight level.
2) *The aircraft speed remains constant*. This is a common assumption both in the path-planning field and en-route ATC, therefore speed changes are excluded here to focus on heading-based conflict resolution.
3) *Ignore aircraft turning dynamics and trajectory uncertainties*. Turning dynamics and trajectory uncertainties significantly complicate the problem. Rather than attempting to model these uncertainties explicitly, we follow established ATC practice: ATCos typically address such uncertainties by applying additional separation buffers, which can likewise be incorporated into the minimum safe separation constraint (i.e., 5 nmi).

Therefore, path planning for en-route aircraft is a problem of determining a path from a start point $p_s$ to a target point $p_t$ given a 2D space $S \subset \mathbb{R}^2$: $(p_s, p_t, S)$. Static obstacles (for example, prohibited airspace, military zones, etc.) are defined as $O = O_1 \cup O_2 \cup \ldots \cup O_n \subset \mathbb{R}^2$. To further limit the complexity of the problem, we discretize the free space $S_{free} = S \setminus O$ using a grid $G = (V, E)$ where $V$ is a set of vertices, $E \subseteq V \times V$ is a set of edges. Then, the 2D path-planning problem can be rewritten as $(p_s, p_t, G)$.

A *path* is a sequence of points $\pi = \{p_1, p_2, \ldots, p_n\} \subset G$ where $p_1 = p_s$ and $p_n = p_t$. The *cost* of a path $\pi$ is the sum of the flight duration between each successive pair of points: $C(\pi) = \sum_{i=1}^{n-1} \mathtt{dist}(i, i+1)/\|\mathbf{v}\|$ where $\mathtt{dist}(i, i+1)$ is the distance between the points $p_i$ and $p_{i+1}$ and $\|\mathbf{v}\|$ is the aircraft speed. We assume that the paths of other aircraft (dynamic obstacles) are known: $\{\pi^1, \pi^2, \ldots, \pi^k\}$. Thus, the goal of the path planning is to find the time-optimal path between $p_s$ and $p_t$ without conflicts with other aircraft.

In path planning, the total cost of a node is $f(n) = g(n_s, n) + h(n, n_t)$. $g(n_s, n)$ denotes the real cost (flight duration) from the start node $n_s$ to the node $n$ and is initially set to infinity. If $g(n_s, n)$ is less than infinity, it means that a feasible path to the node $n$ has been found. $g(n, n')$ is the real cost from the node $n$ to another node $n'$. $h(n, n_t)$ refers to the estimated cost from the node $n$ to the target node $n_t$, which we assume to be the flight duration of the direct path between $n$ and $n_t$. $h(n, n')$ is the estimated cost from $n$ to $n'$. In the path-planning field, $g(n_s, n)$ is usually written as $g(n)$, while $h(n, n_t)$ is commonly simplified to $h(n)$.

For en-route aircraft path planning, several procedural constraints should be considered:

1) *Minimum separation standards*. En-route aircraft should maintain a horizontal separation of at least 5 nmi or a vertical separation of at least 1000 ft. In this research, we only consider horizontal separation.
2) *Maximum allowable time delay*. A Required Time of Arrival (RTA) is usually defined to ensure the efficiency of trajectory-based operations. The maximum allowable time delay indicates the RTA tolerance, meaning that the candidate paths leading to excessive delays should be discarded.
3) *Turning angle limitation*. Aircraft are unable to make sharp turns due to their maneuverability limitations and passenger comfort, which means that the angle between two path segments should be limited.
4) *Minimum distance between waypoints*. The distance between two waypoints should not be excessively short; otherwise, aircraft may be

**Algorithm 1** Solution space generation.

**Require:** start node $n_s$, target node $n_t$, current node $n$, current heading $\theta$, maximum allowable delay $t_{\max,\text{delay}}$, maximum turning angle $\theta_0$, minimum distance between waypoints $d_{\text{near}}$

1: **function** `generateSolutionSpace`$(n, \theta)$
2: solution space $S \leftarrow \emptyset$
3: remaining time $t_n \leftarrow h(n_s) + t_{\max,\text{delay}} - g(n)$
4: $N_{\text{ellipse}} \leftarrow$ `generateSearchRegion`$(n, n_t, t_n)$
5: $N_{\text{circle}} \leftarrow$ `generateNearDistRegion`$(n, d_{\text{near}})$
6: $AOV \leftarrow [\theta - \theta_0, \theta + \theta_0]$
7: **if** zone-based conflict detection is selected **then**
8: Mark conflict cells using `zoneConflictDetection`$(n, \theta)$
9: **end if**
10: $N_{\text{visible}} \leftarrow$ `shadowcasting`$(n, AOV)$
11: **for each** node $n' \in N_{\text{visible}} \cap N_{\text{ellipse}} \setminus N_{\text{circle}}$ **do**
12: **if** zone-based detection is selected or `conflictDetection`$(n, n') =$ false **then**
13: $S \leftarrow S \cup \{n'\}$
14: **end if**
15: **end for**
16: **return** $S$
17: **end function**

unable to execute turns and might deviate significantly from the intended trajectory.

5) *Maximum number of rerouting waypoints*. ATCos may wish to limit the number of rerouting points for aircraft. This constraint enables ATCos to customize the algorithm to align with their preferences.

## 4. Solution space generation

To tackle the path-planning problem described above, we develop an algorithm named Solution Space Path Planning (SSPP). Basically, SSPP constructs a solution space for the next node based on the current node at each step and applies the A* search [25] to iteratively explore the space until the target point is reached. The core of SSPP lies in generating a solution space to identify all reachable nodes from a given node. To accelerate SSPP, we integrate several techniques into the solution space generation process, including three different conflict detection methods. Combining the solution space with search algorithms also allows us to readily visualize the internal process of the algorithm for supporting ATCo supervision [17].

Algorithm 1 shows the solution space generation during the search, including the constraints in Section 3. The `generateSearchRegion` creates an elliptical boundary according to the current and target nodes and the remaining flight time $t_n$. The nodes outside the elliptical boundary are deemed unreachable from the current node due to the maximum allowable time delay constraint. The `generateNearDistRegion` reflects the minimum distance constraint between waypoints, and the nodes inside the circle $N_{\text{circle}}$ should be discarded. Line 6 converts the turning angle constraint into an Angle of View (AOV). Combined with AOV, Line 10 applies shadowcasting [21,22] to find the nodes visible from the current node. More details are presented in Section 4.1.

The `conflictDetection` in Line 12 determines whether a path segment would encounter any conflict with other aircraft (dynamic obstacles). For the zone-based method, the dynamic obstacles are converted into static conflict cells before the shadowcasting is performed, thereby eliminating the need to conduct conflict detection for individual path segments. The conflict detection methods are introduced in Section 4.2.

### 4.1. *Visibility checks*

Before illustrating the shadowcasting algorithm, we first introduce the concept of visibility checks. If two points are mutually visible, the path connecting them is free from collisions with static obstacles. Line-of-sight checking is a common approach for performing such visibility checks on grids in path planning [23]. Specifically, a line drawing algorithm [66,67] can be applied to identify the grid cells traversed by a path, after which these cells are examined to determine whether any of them is blocked by an obstacle. However, line-of-sight checking may be inefficient for computing the visibility between the current node and its reachable nodes during solution space generation, as nodes near the current node may be examined repeatedly. Therefore, rather than using line of sight, a field-of-view approach, such as shadowcasting [21,22], is implemented in SSPP to improve the efficiency of visibility checks.

The shadowcasting algorithm divides the grid into eight octants. Using a linear transformation, the slope range of each octant can all be converted into $[0, 1]$ (i.e., $[0°, 45°]$). As depicted in Fig. 2, the algorithm iterates column by column from the origin (red dot). The red dashed line represents the column currently being scanned, while the green cells denote the cells that have been identified as visible from the origin. At each column, the scan direction is from the blue line to the orange line. When it encounters a blocked cell (black), the scan range shrinks for the subsequent columns. Sometimes, it may yield multiple scan ranges after a few columns because of an obstacle in the middle. The blocked cells cast shadows (gray) that are invisible from the origin. The algorithm ensures that each cell is examined only once to construct the field of view. Given that the nodes are positioned at the centers of the grid cells, the cells are visible only when their centers fall within the scan range. Since the start points of aircraft may not coincide with cell centers, we adapt the algorithm to accommodate such cases. To incorporate the turning angle constraint into the shadowcasting, we also integrate AOV to constrain the scan range accordingly, as shown in Fig. 3. The scan ranges of the octants within AOV are recalculated, while those outside are excluded from the scan.

### 4.2. *Conflict detection*

Conflict detection is central to solution space generation, and its performance can significantly affect the computational efficiency of the overall process. This section presents three distinct conflict detection methods embedded in the solution space generation.

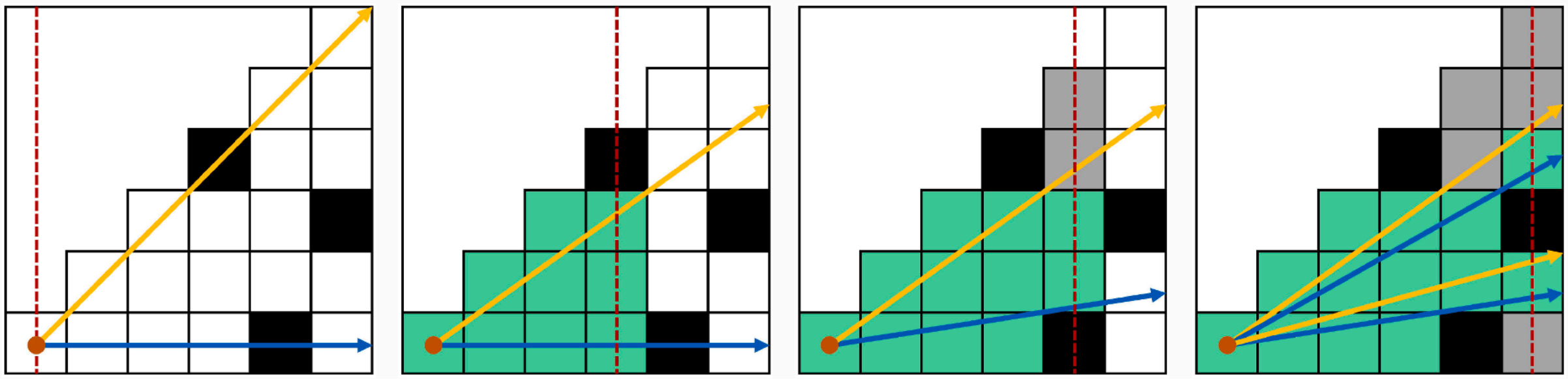

**Fig. 2.** Symmetric recursive shadowcasting, adapted from [22].

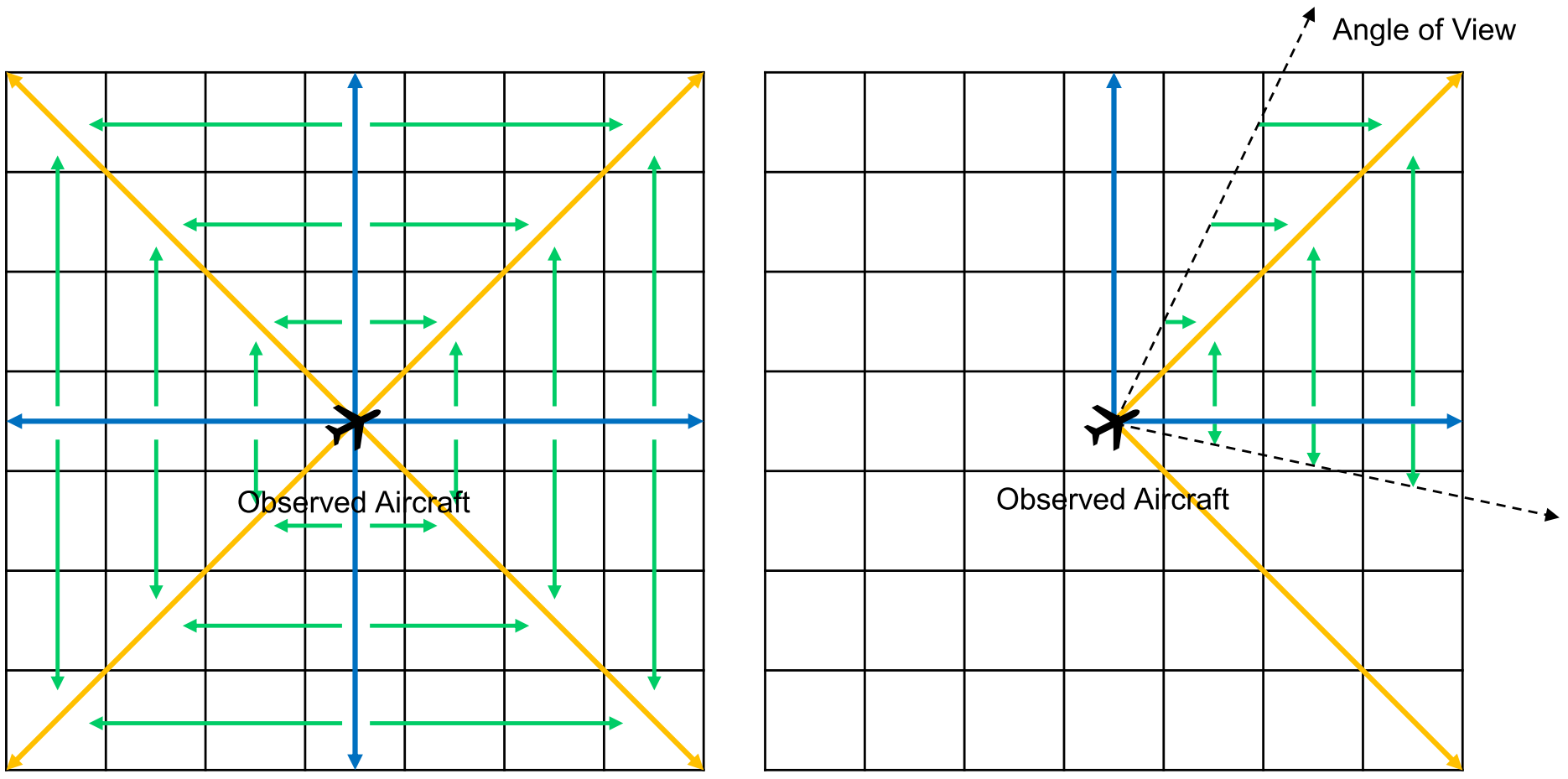

**Fig. 3.** Shadowcasting with angle of view (AOV).

**Algorithm 2** Conflict detection based on separation distance.

1: **function** `distConflictDetection`$(n, n')$
2: path segment $l \leftarrow [n, n']$
3: time interval $T_l \leftarrow [g(n), g(n) + h(n, n')]$
4: **for each** other aircraft $m$ **do**
5: path segments $L_m \leftarrow$ `filterPathSegment`$(L_m, T_l)$
6: **for each** path segment $l_m \in L_m$ **do**
7: $B_l \leftarrow$ `boundingBox`$(l)$
8: $B_{l,m} \leftarrow$ `boundingBox`$(l_m)$
9: $d_x \leftarrow$ `horizontalDist`$(B_l, B_{l,m})$
10: $d_y \leftarrow$ `verticalDist`$(B_l, B_{l,m})$
11: **if** $\max\{d_x, d_y\} < d_{\text{min,sep}}$ **then**
12: $d_{\min} \leftarrow$ `computeMinSeparation`$(l, l_m)$
13: **if** $d_{\min} < d_{\text{min,sep}}$ **then**
14: **return** true
15: **end if**
16: **end if**
17: **end for**
18: **end for**
19: **return** false
20: **end function**

#### 4.2.1. *Conflict detection based on separation distance*

The first method is conflict detection based on separation distance. A conflict is detected if the predicted minimum separation between aircraft is less than the minimum separation standard (i.e., $d_{\text{min,sep}} = 5\,\text{nmi}$). The procedure of this method is shown in Algorithm 2. Lines 5 and 11 indicate that only path segments whose flight time intervals overlap with $T_l$ and that may lead to a loss of separation are considered. To compute the minimum separation between $l$ and $l_m$ in Line 12, we first need to align the start times of $l$ and $l_m$ by trimming the segment that starts earlier. Then, the trimmed segments $l'$ and $l'_m$ can be regarded as rays originating from their respective start points at time $t = 0$. The predicted time to the Closest Point of Approach (CPA) $t_{\text{cpa}}$ and the predicted separation at CPA $\|\Delta\mathbf{p}(t_{\text{cpa}})\|$ can thus be computed by Zou et al. [68]

$$t_{\text{cpa}} = -\frac{\Delta\mathbf{v}^T \Delta\mathbf{p}(0)}{\|\Delta\mathbf{v}\|^2} \quad \text{and} \quad \|\Delta\mathbf{p}(t_{\text{cpa}})\| = \frac{\|\Delta\mathbf{v} \times \Delta\mathbf{p}(0)\|}{\|\Delta\mathbf{v}\|} \tag{1}$$

where $\Delta\mathbf{v}$ is the relative velocity and $\Delta\mathbf{p}(t)$ is the relative position at time $t$, with $\Delta\mathbf{p}(t) = \Delta\mathbf{p}(0) + \Delta\mathbf{v}t$. If $\|\Delta\mathbf{v}\| = 0$, $t_{\text{cpa}}$ is defined as 0. Hence, the predicted minimum separation for two rays is

$$\|\Delta\mathbf{p}\|_{\min} = \begin{cases} \|\Delta\mathbf{p}(t_{\text{cpa}})\| & t_{\text{cpa}} > 0 \\ \|\Delta\mathbf{p}(0)\| & t_{\text{cpa}} \le 0 \end{cases} \tag{2}$$

However, $l'$ and $l'_m$ are not real rays, and the flight time of each path segment should be considered as well. Let $t_l$ and $t_{l,m}$ denote the flight times of $l'$ and $l'_m$, respectively, both measured from the aligned time origin ($t = 0$). Then, $t_{\min} = \min\{t_l, t_{l,m}, t_{\text{cpa}}\}$ and the predicted minimum separation for the path segments is

$$\|\Delta\mathbf{p}\|_{\min} = \begin{cases} \|\Delta\mathbf{p}(t_{\min})\| & t_{\min} > 0 \\ \|\Delta\mathbf{p}(0)\| & t_{\min} \le 0 \end{cases} \tag{3}$$

Eq. (3) can be applied to `computeMinSeparation` in Algorithm 2.

#### 4.2.2. *Conflict detection based on time intervals*

The second conflict detection method is known as grid-based or time-interval-based conflict detection [69,70]. It has been widely applied in the path-planning field, such as Safe Interval Path Planning (SIPP) [50]. The distance-based conflict detection checks whether the distance between aircraft is less than the minimum separation standard within the same period, whereas the time-interval-based approach examines whether the time intervals occupied by aircraft overlap within the same area. As illustrated in Fig. 4, the crossing aircraft is a dynamic obstacle for the observed aircraft, and a conflict arises if there is a temporal overlap between the two aircraft at the intersecting grid cells. The time-interval-based method also eliminates the need to discretize the time dimension into multiple equal-length time slots, resulting in a compact space-time map that effectively reduces the search space.

Algorithm 3 shows the time-interval-based conflict detection method in detail. $C$ is the set of grid cells occupied by crossing aircraft, $I$ denotes the spatiotemporal risk grid map, and $I(c)$ is the union of the occupancy time intervals of all crossing aircraft in cell $c$. Lines 3–8 preprocess the path segments of the crossing aircraft prior to the search and merge their occupancy time intervals into the risk map $I$. Line 13 identifies the cells traversed by the candidate path segment $l$. A conflict is then detected if a temporal overlap occurs in any of these cells.

For the observed aircraft, a line drawing algorithm [66,67] can be adapted to identify occupied cells and compute time intervals. However, for the crossing aircraft (dynamic obstacle), the occupancy computation also needs to account for the minimum separation circle. As shown in Fig. 5, we first identify the grid cells swept by the separation circle and then find the moments when the moving circle enters and leaves each cell. Since the cells are regular squares, it is sufficient to consider the times at which their vertices are first and last covered and their tangent points (red dots) are contacted. The occupancy interval of each cell is determined from the union of these time intervals.

The main advantage of this method is that the trajectory information of dynamic obstacles can be reused and only needs to be mapped once. In sequential planning, the previously mapped trajectories are retained

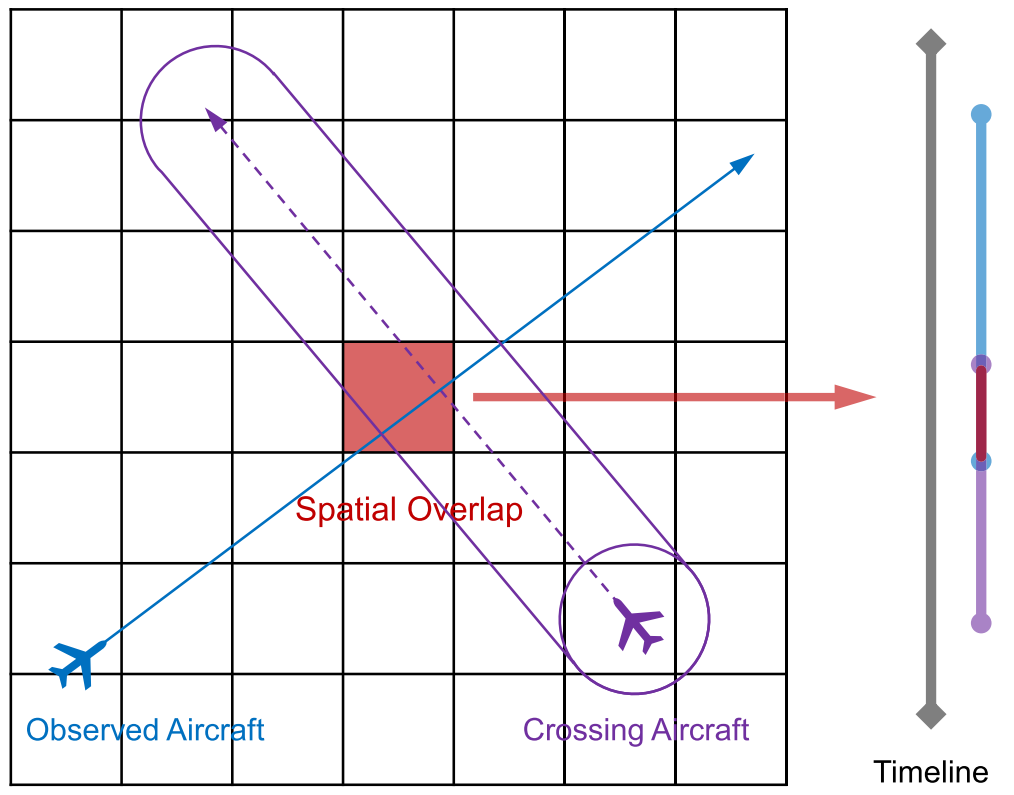

**Fig. 4.** Conflict detection based on time intervals.

**Algorithm 3** Conflict detection based on time intervals.

**function** prepareRiskMap($Grid$)
    $I \leftarrow$ emptyRiskMap($Grid$)
    **for each** other aircraft $m$ **do**
        **for each** path segment $l_m$ **do**
            $I_{l,m} \leftarrow$ mapToGrid($l_m, Grid$)
            $I \leftarrow$ mergeIntervals($I, I_{l,m}$)
        **end for**
    **end for**
    **return** $I$
**end function**

**function** timeConflictDetection($n, n', I$)
    $l \leftarrow [n, n']$
    **for each** cell $c \in$ traverseGrid($l$) **do**
        $I_l(c) \leftarrow$ computeTimeInterval($l, c$)
        **if** $I_l(c) \cap I(c) \neq \emptyset$ **then**
            **return** true
        **end if**
    **end for**
    **return** false
**end function**

**function** mapToGrid($l, Grid$)
    $I_l \leftarrow \emptyset$ and $C_l \leftarrow$ intersectSweptCircle($l, d_{\text{min,sep}}, Grid$)
    **for each** cell $c \in C_l$ **do**
        $I_l(c) \leftarrow$ computeTimeInterval($l, c$)
    **end for**
    **return** $I_l$
**end function**

in the risk map and remain available for subsequent planning. This feature makes the time-interval-based method particularly suitable for scenarios involving a large number of aircraft. However, the accuracy and performance of this method are strongly influenced by the grid size. If the grid size is excessively large, false alarms may occur frequently. A flexible grid size may be helpful in practice.

#### 4.2.3. *Conflict detection based on conflict zones*

The first two conflict detection methods described above are analogous to line-of-sight checks, but for dynamic obstacles. To construct the solution space, each grid cell must be checked for conflicts to determine whether the path connecting it to the current node is obstructed by any dynamic obstacle. However, similar to line-of-sight checks, the nodes around the current node may be examined repeatedly. This redundancy suggests that it would also be beneficial to establish a field-of-view method for dynamic obstacles.

**Algorithm 4** Conflict detection based on conflict zones.

1: **function** zoneConflictDetection($n, \theta$)
2:     $C \leftarrow \emptyset$ and $AOV(n) \leftarrow [\theta - \theta_0, \theta + \theta_0]$
3:     remaining time $t_n \leftarrow h(n_s) + t_{\text{max,delay}} - g(n)$
4:     **for each** other aircraft $m$ **do**
5:         **for each** path segment $l_m$ **do**
6:             $T_{l,m} \leftarrow$ computeFlightTimeInterval($l_m$) $\cap [g(n), g(n) + t_n]$
7:             **if** $T_{l,m} = \emptyset$ or excludeSegment($n, l_m, T_{l,m}$) **then**
8:                 **continue**
9:             **end if**
10:             $\Theta \leftarrow AOV(n) \cap$ computeConflictHeadingRanges($n, l_m$)
11:             $P \leftarrow$ approximateFrontCurve($n, l_m, \Theta, T_{l,m}$)
12:             $C \leftarrow C \cup$ rasterizeFrontCurve($P, Grid$)
13:         **end for**
14:     **end for**
15:     **return** $C$
16: **end function**

Inspired by the Highly Interactive Problem Solver (HIPS) [34], conflict zones can be constructed to identify where conflicts may arise if the observed aircraft only changes its heading, as shown in Fig. 6. The path can be viewed as a ray cast from the current location and the conflict zone can be treated as a static obstacle. Unlike the original HIPS, we incorporate the turning angle constraint (AOV) and the flight time interval into the conflict zone computation. In Fig. 6, the computation is restricted to the yellow region rather than the entire airspace. Consequently, only the red portion of the original conflict zone is retained. The front curve of this limited conflict zone can then be mapped to generate obstacle grid cells, after which shadowcasting is applied to construct the field of view for dynamic obstacles.

In summary, the zone-based method consists of two main steps. First, for the current node, dynamic obstacles are converted into static obstacles using conflict zones. Second, shadowcasting is performed to identify all conflict-free nodes and generate the solution space. This method can be integrated with the visibility checks described in Section 4.1, such that shadowcasting only needs to be performed once for each current node.

Algorithm 4 describes the procedure of conflict zone computation. The empty flight time interval $T_{l,m} = \emptyset$ in Line 7 means that the path segment $l_m$ has already been traversed by the crossing aircraft by the time the observed aircraft reaches node $n$, or that the crossing aircraft will not enter $l_m$ before the latest flight time permitted by the maximum delay constraint. The excludeSegment further skips $l_m$ when a conflict is impossible based on distance. Specifically, $l_m$ is excluded if the observed aircraft cannot reach its vicinity before the crossing aircraft leaves it, or has already passed its vicinity when the crossing aircraft enters it, accounting for the minimum separation.

The conflict heading range $\Theta$ represents the extent of the conflict zone. Instead of considering the entire AOV, we focus only on headings that may lead to a conflict. The conflict heading range can be computed based on the velocity obstacles in the Solution Space Diagram (SSD) [38,71], as presented in Fig. 7. The circle surrounding the crossing aircraft represents the minimum separation, while the circle around the observed aircraft has a radius of $v_o$. Here, $v_o$ and $v_c$ denote the velocities of the observed and crossing aircraft, respectively. Headings outside the velocity obstacle can be excluded from the conflict zone computation. Note that a velocity obstacle may produce two separate conflict heading ranges due to the four intersections between the obstacle triangle and the velocity circle. For each heading $\theta_i$ within these ranges, the time interval and position of the potential conflict can be computed. As depicted in Fig. 8, if the start time $t_0$ of the crossing aircraft is later than $g(n)$, its initial position can be back-propagated to align the two start times, thereby facilitating the computation of the conflict time interval.

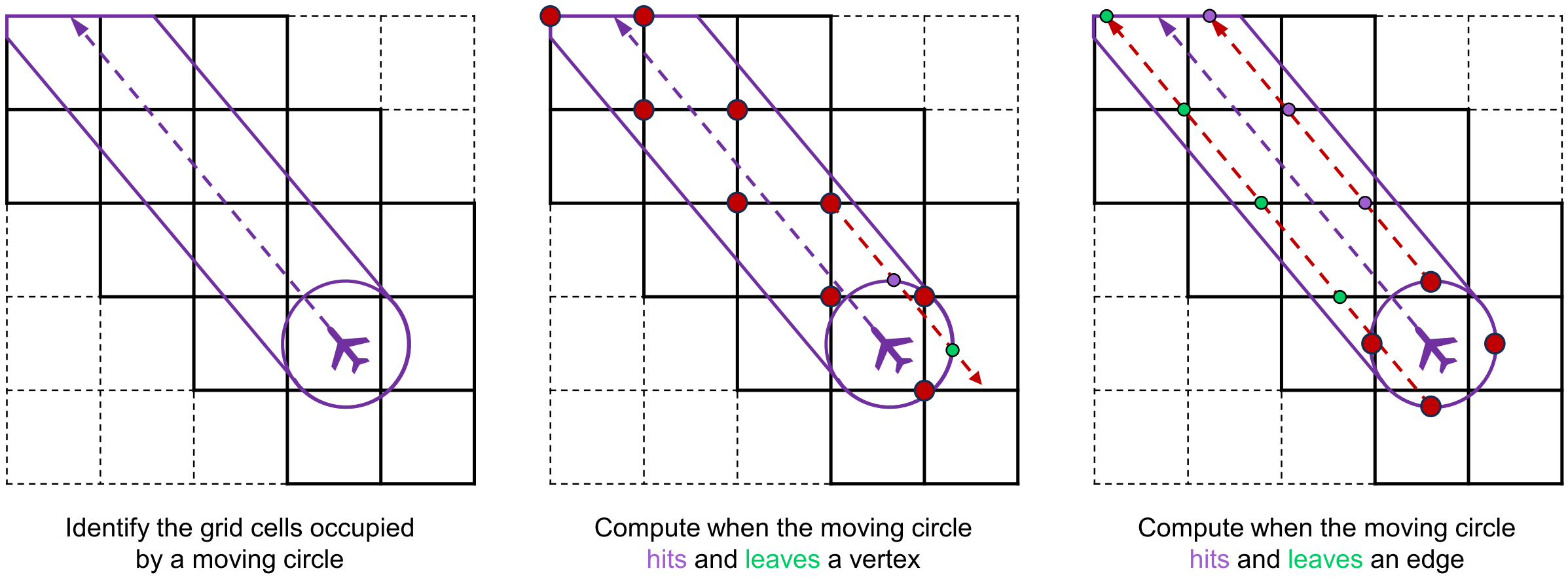

**Fig. 5.** Computation of occupancy time intervals (`mapToGrid`).

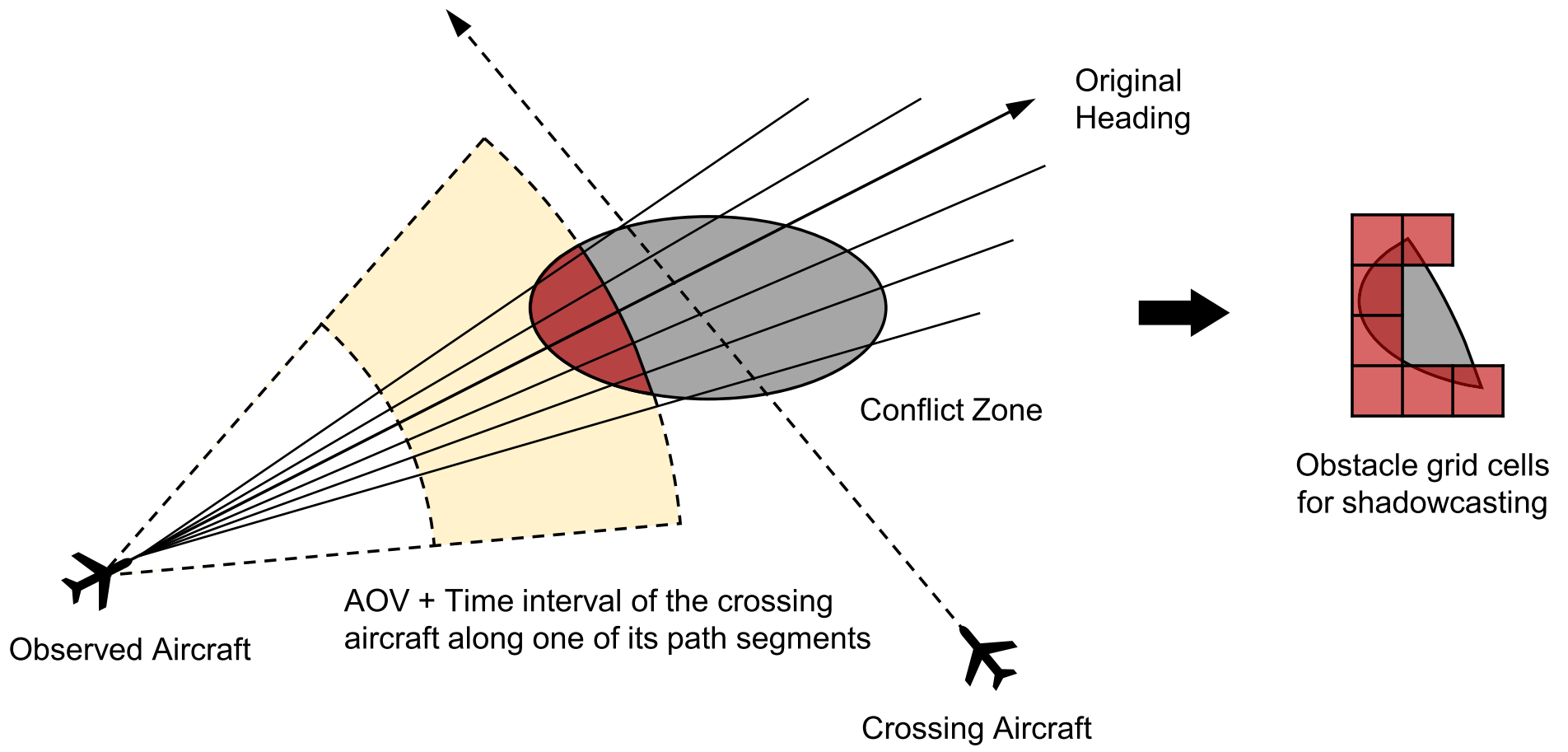

**Fig. 6.** Construction of a conflict zone, combined with AOV and flight time interval.

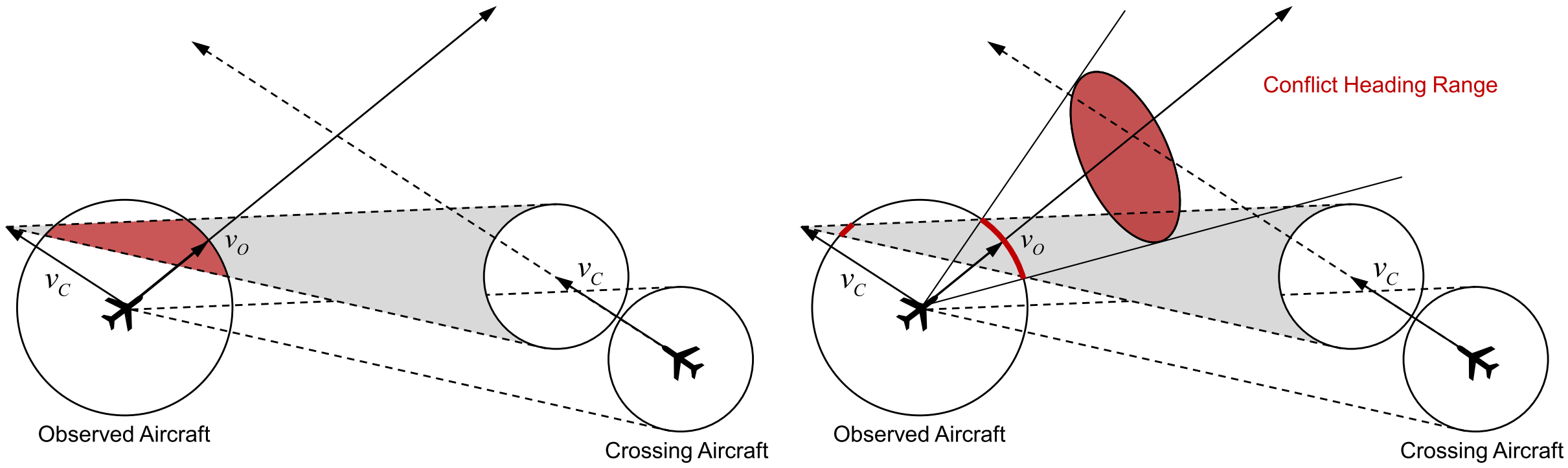

**Fig. 7.** Computation of conflict heading ranges based on the triangular velocity obstacle in SSD.

A straightforward approach to computing the front curve of the conflict zone is to discretize the headings of the observed aircraft within the conflict heading ranges. However, with a fixed angular resolution, the spacing between adjacent front points increases with distance from the current node, reducing the approximation accuracy. Therefore, we first derive an analytical expression for the conflict zone boundary and then approximate its front curve using an *adaptive piecewise-linear* method based on tangents and chords, rather than relying on fixed heading discretization.

Following the time alignment, we set $t = 0$ at departure from $\mathbf{p}_n$, the position of node $n$ (the absolute time is $g(n)$). Let $\mathbf{p}_{l,m}(0)$ denote the virtual starting position of the crossing aircraft $m$, obtained by propagating its motion along $l_m$ backward or forward to the aligned time origin. The initial relative position is $\Delta\mathbf{p}(0) = \mathbf{p}_n - \mathbf{p}_{l,m}(0)$.

The flight time interval $T_{l,m}$ is expressed relative to $g(n)$ as $[t_a, t_b]$. The propagated motion is used only to compute relative positions, while conflicts are evaluated only for $t \in [t_a, t_b]$. For a heading $\theta_i \in \Theta$, we define the unit direction vector $\mathbf{u}(\theta_i) = (\cos\theta_i, \sin\theta_i)^T$. The relative velocity is $\Delta\mathbf{v}(\theta_i) = \|\mathbf{v}\|\mathbf{u}(\theta_i) - \mathbf{v}_{l,m}$, where $\|\mathbf{v}\|$ is the speed of the observed aircraft and $\mathbf{v}_{l,m}$ the velocity of the crossing aircraft along $l_m$. The minimum separation boundary is reached when

$$\|\Delta\mathbf{p}(0) + \Delta\mathbf{v}(\theta_i)t\|^2 - d^2_{\text{min,sep}} = at^2 + 2bt + c = 0 \quad (4)$$

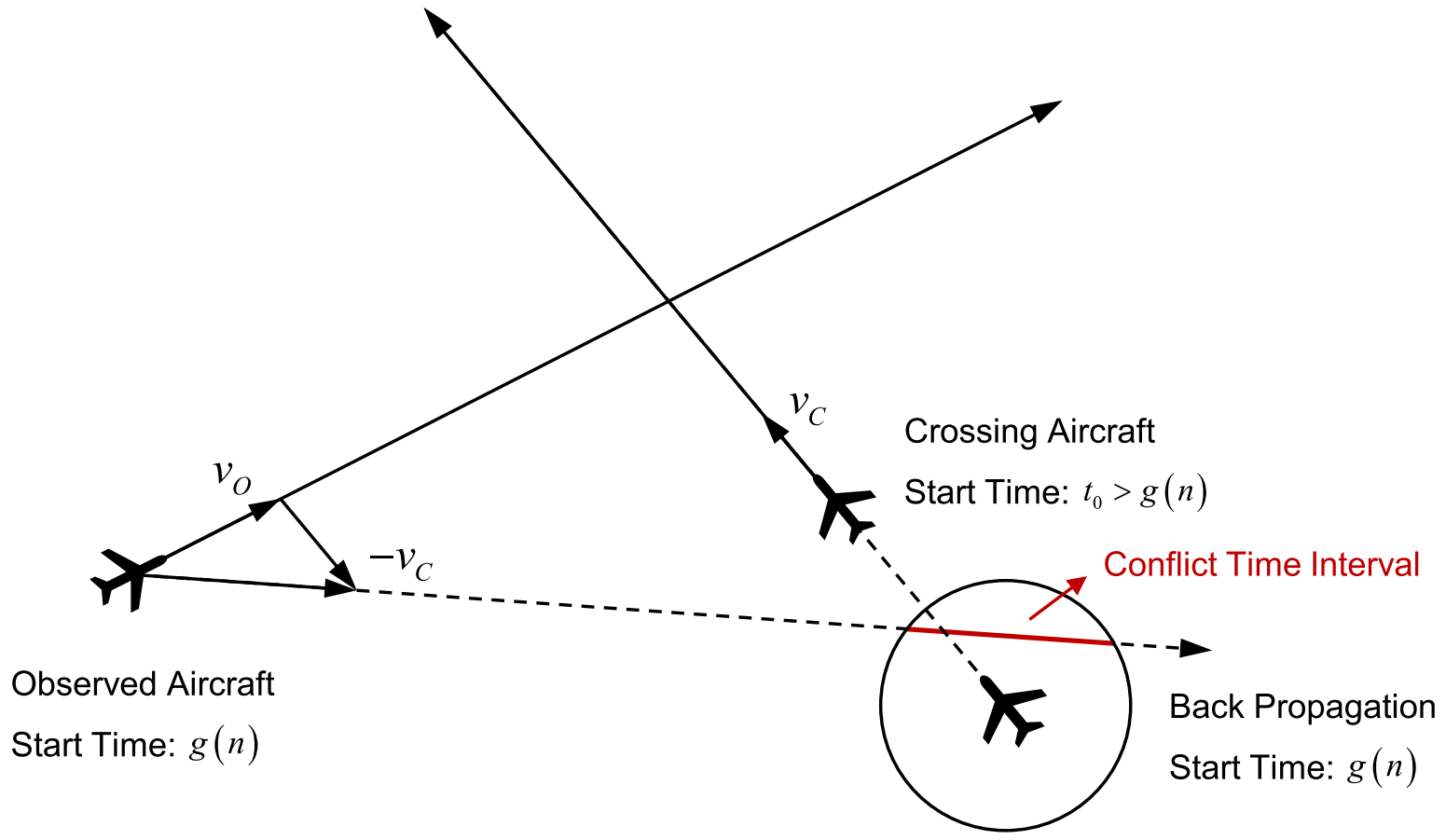

**Fig. 8.** Computation of conflict time interval.

where $a = \|\Delta\mathbf{v}(\theta_i)\|^2$, $b = \Delta\mathbf{p}(0)^T\Delta\mathbf{v}(\theta_i)$ and $c = \|\Delta\mathbf{p}(0)\|^2 - d^2_{\text{min,sep}}$. For $a > 0$ and $\Delta = b^2 - ac \geq 0$, the two roots

$$t_-(\theta_i) = \frac{-b - \sqrt{\Delta}}{a}, \quad t_+(\theta_i) = \frac{-b + \sqrt{\Delta}}{a} \tag{5}$$

give the entry and exit times of the minimum separation circle. Their corresponding positions constitute the original conflict zone boundary. The time-constrained front curve is formed by the first conflict position along each heading $\theta_i$ within $[t_a, t_b]$. Its time is therefore $t_f(\theta_i) = \max\{t_a, t_-(\theta_i)\}$, and its position is

$$\mathbf{p}_f(\theta_i) = \mathbf{p}_n + \|\mathbf{v}\| t_f(\theta_i)\mathbf{u}(\theta_i) \tag{6}$$

A front point is valid only if $t_f(\theta_i) \leq \min\{t_b, t_+(\theta_i)\}$. When $\Delta < 0$, no conflict occurs along that heading. If $a = 0$, the relative position remains constant, and the entire interval is either in conflict or conflict-free depending on whether the initial separation is within the minimum separation.

Eq. (6) indicates two forms of front curves. When $t_-(\theta_i) > t_a$, the front follows the original conflict boundary. When $t_-(\theta_i) \leq t_a \leq t_+(\theta_i)$, the aircraft are already within the minimum separation at the start of the valid flight interval, and the front follows a circular arc centered at node $n$ with radius $\|\mathbf{v}\|t_a$. Therefore, the conflict heading range $\Theta$ can be further divided into subranges corresponding to the original front, the fixed-time arc, and headings with no conflict.

For an original front, we approximate the curve using the tangents at its endpoints. To derive the tangent direction, define $\mathbf{u}_\perp(\theta_i) = (-\sin\theta_i, \cos\theta_i)^T$ as the unit vector perpendicular to $\mathbf{u}$ in the direction of increasing heading, and let

$$\kappa = (\Delta\mathbf{p}(0) - \mathbf{v}_{l,m}t_-(\theta_i))^T\mathbf{u}_\perp(\theta_i) \tag{7}$$

Here, $\Delta\mathbf{p}(0) - \mathbf{v}_{l,m}t_-$ is the vector from the position of the crossing aircraft at $t_-$ to node $n$, and $\kappa$ is its signed projection onto $\mathbf{u}_\perp(\theta_i)$. For $\Delta > 0$, differentiating Eq. (4) with respect to $\theta_i$ along the entry boundary yields

$$\frac{dt_-}{d\theta_i} = \frac{\kappa\|\mathbf{v}\|t_-(\theta_i)}{\sqrt{\Delta}} \tag{8}$$

Using Eq. (8) in the derivative of Eq. (6) gives the tangent vector

$$\frac{d\mathbf{p}_f}{d\theta_i} = \frac{\|\mathbf{v}\|t_-(\theta_i)}{\sqrt{\Delta}}\mathbf{d}(\theta_i), \quad \mathbf{d}(\theta_i) = \kappa\|\mathbf{v}\|\mathbf{u}(\theta_i) + \sqrt{\Delta}\mathbf{u}_\perp(\theta_i). \tag{9}$$

Hence, $\mathbf{d}(\theta_i)$ provides the tangent direction of the original front.

Consider a heading range $[\theta_a, \theta_b]$ corresponding to the original front, with midpoint heading $\theta_c = (\theta_a + \theta_b)/2$. Let $\mathbf{p}_a$, $\mathbf{p}_b$, and $\mathbf{p}_c$ denote the front positions at these headings, respectively, and let $\mathbf{d}_a$ and $\mathbf{d}_b$ denote the tangent directions at the endpoints. If the two tangent lines are not parallel, their intersection is

$$\mathbf{q} = \mathbf{p}_a + \frac{(\mathbf{p}_b - \mathbf{p}_a)\times\mathbf{d}_b}{\mathbf{d}_a\times\mathbf{d}_b}\mathbf{d}_a, \tag{10}$$

where $\times$ denotes the two-dimensional scalar cross product, and $\mathbf{a}\times\mathbf{b} = a_x b_y - a_y b_x$. The two line segments $[\mathbf{p}_a, \mathbf{q}]$ and $[\mathbf{q}, \mathbf{p}_b]$ provide a piecewise-linear approximation of this portion of the front. To determine whether further subdivision is required, we evaluate the sampled front length $L$ and the distance $e$ from $\mathbf{p}_c$ to $[\mathbf{p}_a, \mathbf{q}]$ and $[\mathbf{q}, \mathbf{p}_b]$:

$$\begin{aligned} L &= \|\mathbf{p}_c - \mathbf{p}_a\| + \|\mathbf{p}_b - \mathbf{p}_c\| \\ e &= \min\{\texttt{dist}(\mathbf{p}_c, [\mathbf{p}_a, \mathbf{q}]), \texttt{dist}(\mathbf{p}_c, [\mathbf{q}, \mathbf{p}_b])\} \end{aligned} \tag{11}$$

Here, `dist` denotes the shortest distance from a point to a finite line segment. The tangent approximation is accepted when $L \leq d_{\text{grid}}$ and $e \leq d_{\text{grid}}/16$, where $d_{\text{grid}}$ denotes the grid size. Otherwise, the heading range is bisected at $\theta_c$, and each subrange is evaluated recursively. These criteria allow the heading resolution to adapt to the local geometry and spatial scale of the front curve. Consequently, the shape of distant conflict zones can still be captured accurately, even when they are far from node $n$.

If the endpoint tangents are collinear, the endpoint chord is used instead. The chord is also used when a tangent-based approximation cannot be constructed or further subdivision is not possible. Fixed-time arcs, in contrast to the original front, are approximated directly by chords whose lengths do not exceed $0.5d_{\text{grid}}$ in this study. Different from the tangent-based approximation, these chords lie closer to node $n$ than the corresponding arcs.

The line segments obtained from the tangent- and chord-based approximations together form the piecewise-linear front approximation $P$ returned by `approximateFrontCurve` in Algorithm 4. Finally, `rasterizeFrontCurve` clips these segments to the grid and marks all intersected cells. The union of these cells, denoted by $C$, is then used as the dynamic obstacle input for the subsequent shadowcasting scan.

This zone-based approach eliminates the need to perform conflict detection for each grid cell and is readily compatible with visibility checks such as shadowcasting. However, the piecewise-linear approximation of the front curve may introduce some loss of precision, which is controlled by the spatial subdivision criteria described above.

## 5. Solution space path planning

In Section 4, we introduced the details of solution space generation at each search step, and, in this section, we will illustrate the main procedures and algorithms for Solution Space Path Planning (SSPP). Two versions of SSPP are proposed: *SSPPV and SSPPE*. They differ in how search nodes are defined: SSPPV treats each vertex/point (i.e., a grid cell) as a node, whereas SSPPE treats each ordered pair of vertices/points (i.e., a directed edge) as a node. We also explore a potential extension to SSPP that accounts for robustness by jointly optimizing delay and separation.

**Algorithm 5** Vertex-based solution space path planning (SSPPV).

```
Require: start node n_s, target node n_t, heading θ_s
1: open ← {n_s}, closed ← ∅, g(n_s) ← 0, depth(n_s) ← 0
2: while open ≠ ∅ do
3:     n ← arg min_{n∈open} f(n)
4:     open ← open \ {n}, closed ← closed ∪ {n}
5:     if n = n_t then
6:         return pathTo(n)
7:     end if
8:     if depth(n) > k then
9:         continue
10:    end if
11:    θ ← computeHeading(parent(n), n) or θ_s
12:    S(n) ← generateSolutionSpace(n, θ)
13:    for each node n′ ∈ S(n) do
14:        if n′ ∈ closed then
15:            continue
16:        end if
17:        g_new ← g(n) + g(n, n′)
18:        if parent(n′) = null or g(n′) > g_new then
19:            parent(n′) ← n, g(n′) ← g_new
20:            f(n′) ← g_new + h(n′)
21:            depth(n′) ← depth(n) + 1
22:            insert or update n′ in open
23:        end if
24:    end for
25: end while
26: return null
```

**Algorithm 6** Edge-based solution space path planning (SSPPE).

```
Require: start node n_s, target node n_t, heading θ_s
l_s ← [n_s, n_s], g(l_s) ← 0, depth(l_s) ← 0
open ← {l_s}, closed ← ∅
while open ≠ ∅ do
    l = [n_1, n_2] ← arg min_{l∈open} f(l)
    open ← open \ {l}, closed ← closed ∪ {l}
    if n_2 = n_t then
        return pathTo(l)
    end if
    if depth(l) > k then
        continue
    end if
    θ ← computeHeading(n_1, n_2) or θ_s
    S(l) ← generateSolutionSpace(n_2, θ)
    for each l′ = [n_2, n′] ∈ S(l) do
        if l′ ∈ closed then
            continue
        end if
        g_new ← g(l) + g(n_2, n′)
        if parent(l′) = null or g(l′) > g_new then
            parent(l′) ← l, g(l′) ← g_new
            f(l′) ← g_new + h(n′)
            depth(l′) ← depth(l) + 1
            insert or update l′ in open
        end if
    end for
end while
return null
```

### 5.1. SSPPV: vertex-based SSPP

The implementation of SSPPV is straightforward: it replaces the eight neighboring nodes considered in traditional A* with a solution space during node expansion, as shown in Algorithm 5. The solution space $S(n)$ can be regarded as extended neighbors and represents all reachable nodes from the current node $n$. This approach enables any-angle path planning [23], which allows any-angle turns at graph vertices to generate straighter paths without post-hoc smoothing. This property is particularly suitable for aircraft that cannot make sharp turns. Certainly, the turning angle constraint should be satisfied and $S(n)$ inherently incorporates this requirement (AOV). We also introduce $depth(n)$ to indicate the depth of a node, allowing ATCos to restrict the depth of the search tree based on the maximum number of rerouting points $k$ ($k \geq 0$). Since the nodes in $S(n)$ are reachable from the current node $n$, it is possible for $n$ to be the true or best parent of a node $n'$ in $S(n)$. From Line 13 to Line 24, a comparison is conducted between the current node $n$ and the current best parent of $n'$ to determine which node is a better parent for $n'$. This operation is intended to expand and rewire the search tree to achieve a growing and straight tree structure. The closed nodes are excluded in $S(n)$ because, once a node is closed, the algorithm retains its current best-known path and does not consider alternative paths.

The solution space generated at each search step appears to cover all possibilities that may contribute to finding an optimal path during the search. However, this is not necessarily the case because the generated solution space also depends on the "incoming" heading $\theta$. Due to the AOV constraint, different incoming headings may lead to different sets of feasible outgoing headings. Consequently, a node $n'$ may be unreachable from a closed node $n$ and its parent $parent(n)$, but reachable through $n$ and another parent $parent'(n)$. Unlike traditional A*, node expansion in SSPPV therefore depends not only on the node position but also on its parent. The closed nodes and their parents form a fixed search tree, from which `generateSolutionSpace`$(n, \theta)$ further explores feasible expansions.

Therefore, SSPPV is not complete: it may fail to find a feasible path because an alternative incoming heading at a previously closed node is not retained. It thus does not provide an optimality guarantee either.

### 5.2. SSPPE: edge-based SSPP

To address this limitation, we develop SSPPE, as shown in Algorithm 6. Since node expansion in SSPPV depends on both nodes and their parents, SSPPE treats a pair of points (directed edge), $l = [n_1, n_2]$, as a search node. Each node therefore corresponds to a specific incoming heading at $n_2$. The real cost of a node $l$ is $g(l) = g(parent(l)) + g(n_1, n_2)$ and the total cost is $f(l) = g(l) + h(n_2)$. Once $l$ is closed, the lowest-cost path to $l$ is considered to have been found. However, this approach extends the search space from $\mathbb{R}^2$ to $\mathbb{R}^2 \times \mathbb{R}^2$, as each node comprises a pair of points. SSPPE may thus be significantly slower than SSPPV.

However, SSPPE is still neither complete nor optimal because retaining only the earliest arrival at a directed-edge state does not guarantee that conflicts with dynamic obstacles along subsequent paths can be resolved. For example, as shown in Fig. 9, consider a case where point $C$ is unreachable from node $AB$ (a directed edge) when the arrival time at $B$ is 13:00, but becomes reachable if the arrival time is delayed to 13:10. In such situations, the earliest arrival time does not produce a *complete* solution space that represents all possibilities from $AB$, indicating that SSPPE is not optimal. Interestingly, compared to TO-AA-SIPP, optimal any-angle path planning becomes more challenging when waiting actions are not permitted.

Both SSPPV and SSPPE are based on A* search but differ in their state representations. Let $N$ denote the number of traversable nodes. SSPPV has at most $N$ states, while SSPPE has at most $N^2$ directed-edge states. In the worst case, each state in either algorithm has $O(N)$ candidate successors. Since closed states are not reopened, the total numbers of candidate successor checks are bounded by $O(N^2)$ for SSPPV and $O(N^3)$ for SSPPE, excluding the costs of visibility checks, conflict detection, and priority-queue operations. Thus, the additional computational cost of SSPPE primarily stems from its larger state space.

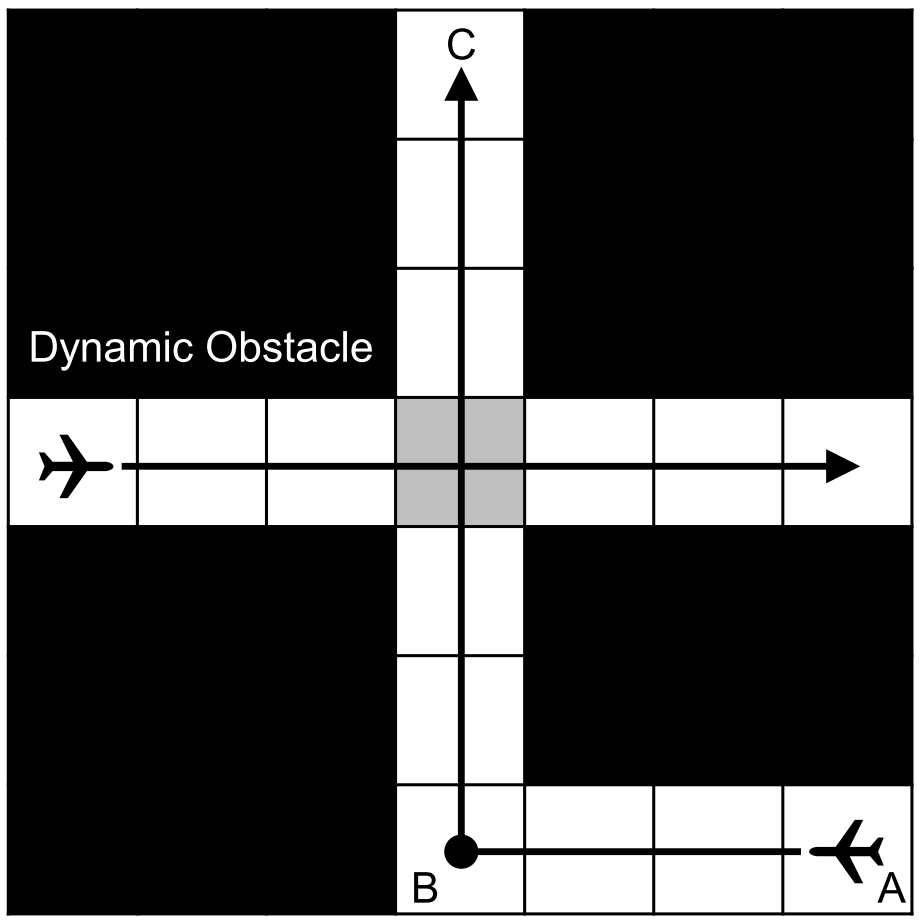

**Fig. 9.** Example illustrating why SSPPE is not complete and therefore does not guarantee optimality.

### 5.3. Trade-off between delay and separation

The original objective of SSPP is to minimize flight time (or delay). This means that only the node with the lowest delay is selected from *open* and moved to *closed* at each step. However, in practice, ATCos may favor more robust solutions [37]. In this section, we explore potential extensions to SSPP by taking robustness into account.

Idris et al. [72,73] introduced the concept of trajectory flexibility and proposed two metrics for it: robustness and adaptability. Klomp et al. [37] further developed metrics for point-based, trajectory-based, sector-based and control-based robustness. However, calculating these metrics often requires generating a complete “trajectory” solution space, which typically demands significant computing resources and time. While SSPP computes the “next-point” solution space for each closed node, the “trajectory” solution space it explores remains limited.

Since disturbances may arise from the maneuvering behavior or uncertainties of aircraft, robustness can be measured by the separation distance between them. A larger separation indicates a greater ability to tolerate such disturbances. Similarly, adaptability can be evaluated by the remaining allowable time delay. The time-optimal trajectory suggests that the aircraft has the maximum available maneuvering space to adjust its current flight plan. Therefore, we can redesign the cost function of SSPP to minimize the time delay and maximize the separation distance:

$$f_{\text{total}}(n) = g_{\text{total}}(n) = w_1 \cdot g_{\text{delay}}(n) + w_2 \cdot g_{\text{sep}}(n) = w_1 \cdot \frac{t_{\text{delay}}(n)}{t_{\text{max,delay}}} + (1-w_1) \cdot \frac{d_{\text{min,sep}}}{d_{\text{sep}}(n)} \tag{12}$$

where $w_1$ and $w_2$ are the objective weights within $[0,1]$, $t_{\text{delay}}(n)$ is the estimated arrival delay at the target via node $n$ and $t_{\text{max,delay}}$ is the maximum allowable time delay. $t_{\text{delay}}(n)$ can be calculated by

$$t_{\text{delay}}(n) = g(n) + h(n) - h(n_s) = f(n) - h(n_s) \tag{13}$$

where $n_s$ is the start node, $g(n)$ is the real flight time from the start node to node $n$ and $h(n)$ is the estimated flight time from node $n$ to the target node. Here, $f(n)$ in Eq. (13) denotes the original unweighted cost $g(n) + h(n)$, rather than the weighted priority. $d_{\text{sep}}(n)$ is the separation distance between `pathTo`$(n)$ and other crossing aircraft and $d_{\text{min,sep}}$ is the minimum separation. $d_{\text{sep}}(n)$ can be calculated by

$$d_{\text{sep}}(n) = \min\{d_{\text{sep}}(parent(n)), d_{\text{sep}}(parent(n), n)\} \tag{14}$$

where $d_{\text{sep}}(parent(n), n)$ is the separation between the path segment $[parent(n), n]$ and other aircraft. The advantage of this design is that $g_{\text{delay}}(n)$ and $g_{\text{sep}}(n)$ are both scaled to $[0,1]$, eliminating the need to account for units. Moreover, both $g_{\text{delay}}(n)$ and $g_{\text{sep}}(n)$ are *monotonic* for a trajectory; they never decrease as the depth of the waypoints increases.

For robust SSPPV, cost computation and comparison in Algorithm 5 require modifications:

$$g_{\text{new}} \leftarrow w_1 \cdot \frac{g(n) + g(n,n') + h(n') - h(n_s)}{t_{\text{max,delay}}} + (1-w_1) \cdot \frac{d_{\text{min,sep}}}{\min\{d_{\text{sep}}(n), d_{\text{sep}}(n,n')\}} \tag{15}$$

$f(n') \leftarrow g_{\text{new}}$, $g(n')$ is maintained separately as the accumulated flight time, and candidate paths are compared based on their weighted costs. Similarly, in robust SSPPE,

$$g_{\text{new}} \leftarrow w_1 \cdot \frac{g(l) + g(n_2,n') + h(n') - h(n_s)}{t_{\text{max,delay}}} + (1-w_1) \cdot \frac{d_{\text{min,sep}}}{\min\{d_{\text{sep}}(l), d_{\text{sep}}(n_2,n')\}} \tag{16}$$

where $l = [n_1, n_2]$ and $f(l') \leftarrow g_{\text{new}}$. According to Eq. (14),

$$d_{\text{sep}}(l) = \min\{d_{\text{sep}}(parent(l)), d_{\text{sep}}(n_1,n_2)\} \tag{17}$$

If $w_1 = 1$, robust SSPP reduces to its basic version. This is because $t_{\text{max,delay}}$ is a positive constant and $f_{\text{total}}(n) = g_{\text{delay}}(n)$ increases monotonically with $f(n)$. In this case, the separation distance may approach the minimum separation as the algorithm prioritizes trajectories with minimal delay. If $w_1 = 0$, robust SSPP seeks paths with the largest possible separation, even though the resulting delay may approach the maximum allowable value. ATCos should be allowed to adjust the weights to balance these two objectives based on their operational needs.

Since $d_{\text{sep}}(n)$ involves computing the separation distance, it is ideally suited for the distance-based conflict detection. For the other two methods, alternative cost function designs may be required. In this article, we only explore one of the possibilities for extensions to SSPP. Essentially, the cost function affects which node is selected from the solution space $S(n)$. A more complex function incorporating additional factors can also be developed.

## 6. Case study

### 6.1. Experiment setup

To analyze the performance of the proposed algorithms, we selected the Delta sector in the Maastricht Upper Area Control Centre (MUAC) airspace above the Netherlands as a case study. A single sector, rather than a multi-sector coordination environment, was chosen because our focus is on tactical ATC rather than air traffic flow management. The algorithms were implemented in a web-based ATC simulator[1] that we developed using JavaScript and the experiments were performed on Node.js v22.19.0 on a laptop with 2.30 GHz Intel Core i7-11800H and 16 GB RAM.

Drawing on our experience in ATC, we set the algorithm parameters as shown in Table 1. The minimum horizontal separation 5 nmi is determined by the minimum separation standard. The maximum allowable time delay is set to 300 seconds because the resulting initial solution space nearly covers the entire sector. The turning angle limitation is set to 60° to reflect typical aircraft maneuverability. The minimum distance between waypoints is set to 10 nmi (twice the minimum separation standard) based on discussions with ATCos. The maximum number of rerouting points is set to either 1 or 5 to evaluate the extent of improvement achievable by allowing additional rerouting points in path planning. The grid size is set to 2.5 nmi, allowing ATCos to more easily observe whether two aircraft are safely separated on grids. Sensitivity analyses are also conducted in this section to evaluate the effects of different parameter settings on algorithm performance.

[1] Demo: https://atclab.tudelft.nl/sspp/.
Code: https://github.com/yiyuanzou/sspp.

**Table 1**
Algorithm parameters.

| Parameter | Value |
|---|---|
| Minimum horizontal separation | 5 nmi |
| Maximum allowable time delay | 300 s |
| Turning angle limitation | 60° |
| Minimum distance between waypoints | 10 nmi |
| Maximum number of rerouting points | 1 or 5 |
| Grid size | 2.5 nmi |

There were about 38 entry and exit points for the Delta sector. We randomly generated pairs of these points to be start and target points for aircraft. By default, the aircraft flew directly to their designated target points. The flight level was set to FL350, with a random speed from Mach 0.71 to Mach 0.83. All aircraft entered the sector simultaneously and executed the proposed path-planning algorithm sequentially. The aircraft that performed the algorithm first were treated as dynamic obstacles for the subsequent aircraft. In practice, we can adopt a first-come-first-served principle. If an aircraft could not find a conflict-free path, it would proceed along its original direct route and be marked as failed. Here, “conflict-free” means that the minimum separation is not violated. To avoid generating unrealistic flights, the following rules were applied:

1) The start points of aircraft should be at least 7.5 nmi apart to ensure that no loss of separation occurs at the beginning. This value was determined based on the minimum separation standard with a 50% buffer.
2) There should be no conflict within 2 min when aircraft follow direct routes to their target points. This value was chosen because the short-term conflict alert in ATC is typically 2 min in advance.
3) The start and target points of aircraft should not be located on the same edge of the sector.
4) To ensure a flight impacts the sector, the length of its direct route was set to exceed 30 nmi.
5) The direct route should be within the sector without intersecting its boundary.

In normal scenarios, there is no static obstacle in the sector. However, if hazardous weather appears, a no-fly zone may need to be established. Therefore, we defined two different airspace conditions: one with no static obstacle and another with a single obstacle located in the sector center. Fig. 10 presents the screenshots of a random scenario with 12 aircraft. The radar labels are blue, indicating that these aircraft are under automation control. The blue rectangular region is a no-fly zone (static obstacle) that all aircraft need to avoid entering. These paths were generated by SSPP based on conflict zones, serving as an example. Fig. 11 visualizes the internal process of this algorithm. The green ellipse represents the maximum allowable delay constraint, incorporating the AOV and minimum waypoint distance constraints. The green cells represent search nodes explored by the algorithm, with light yellow lines representing the branches connecting these nodes. The red regions are conflict zones and the one under the mouse cursor is caused by the aircraft “RN003”. All areas within the green ellipse, except for the green cells, are considered unreachable due to conflicts. To reduce visual clutter, the red cells indicating unreachable areas are hidden. Note that the conflict zones are generated exclusively for the zone-based conflict detection method. The distance-based and time-interval-based methods only specify whether nodes are reachable or not.

### 6.2. Performance analysis

To evaluate the algorithm performance, we first use the parameter settings shown in Table 1 by default, with the maximum number of rerouting points set to 5. Results from scenarios with and without a no-fly zone in the sector are combined. The three conflict detection methods are represented by D (distance-based), T (time-interval-based), and Z (zone-based). Runtime is reported as the average per aircraft and includes both successful and unsuccessful searches. As a human-centered design objective we adopt 100 ms as an aspirational runtime target to provide users with a sense of immediate feedback during frequent what-if interactions, paving the way for uninterrupted, real-time exploration of potential solutions [14]. The success rate is defined as the proportion of aircraft for which a conflict-free path is found. This metric is adopted because planning is performed sequentially rather than simultaneously for all aircraft. Mean delay is calculated only for successful planning cases, while speedup is computed as the mean of paired runtime ratios across samples.

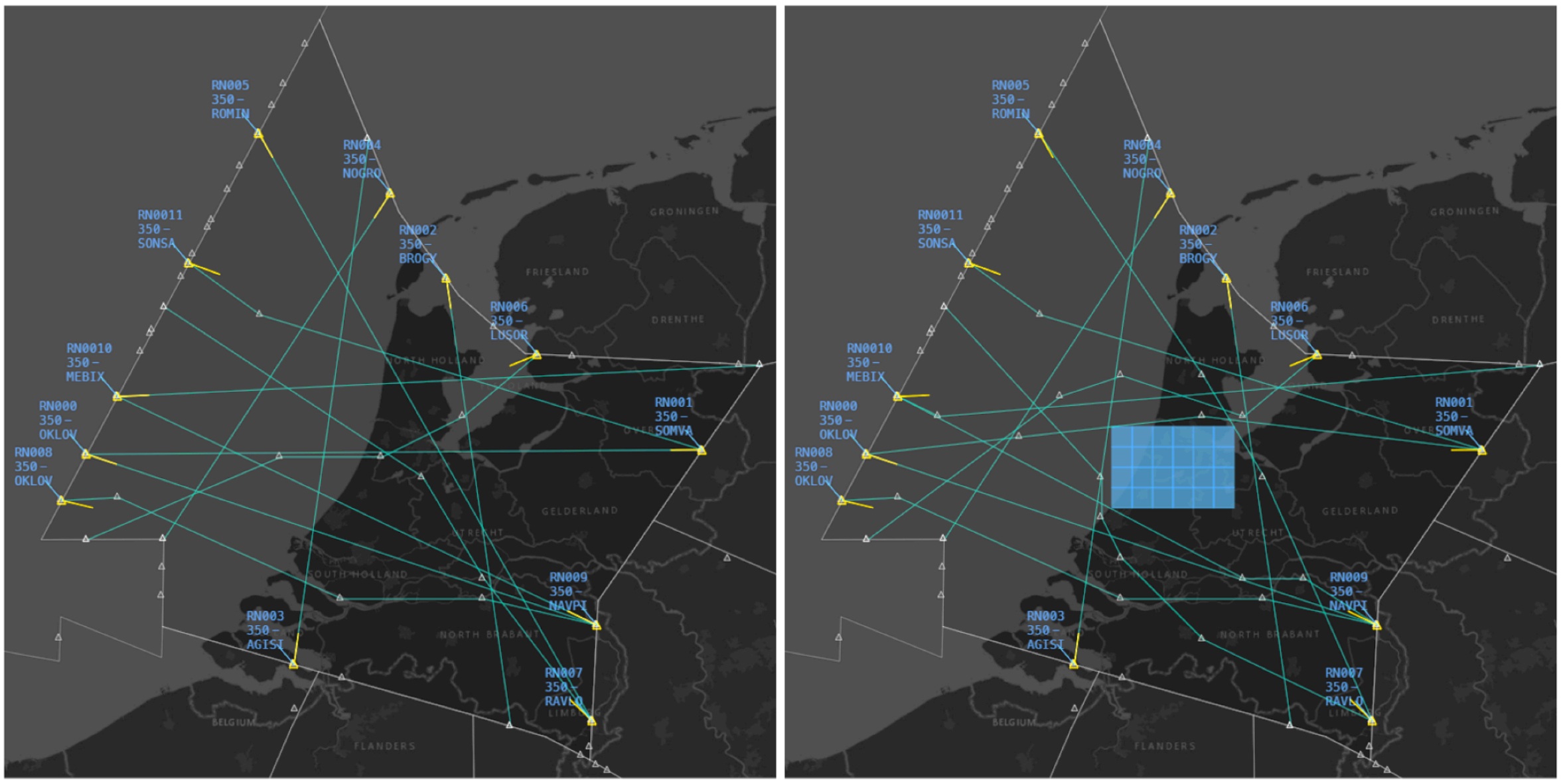

**Fig. 10.** Screenshots of a random scenario featuring 12 aircraft.

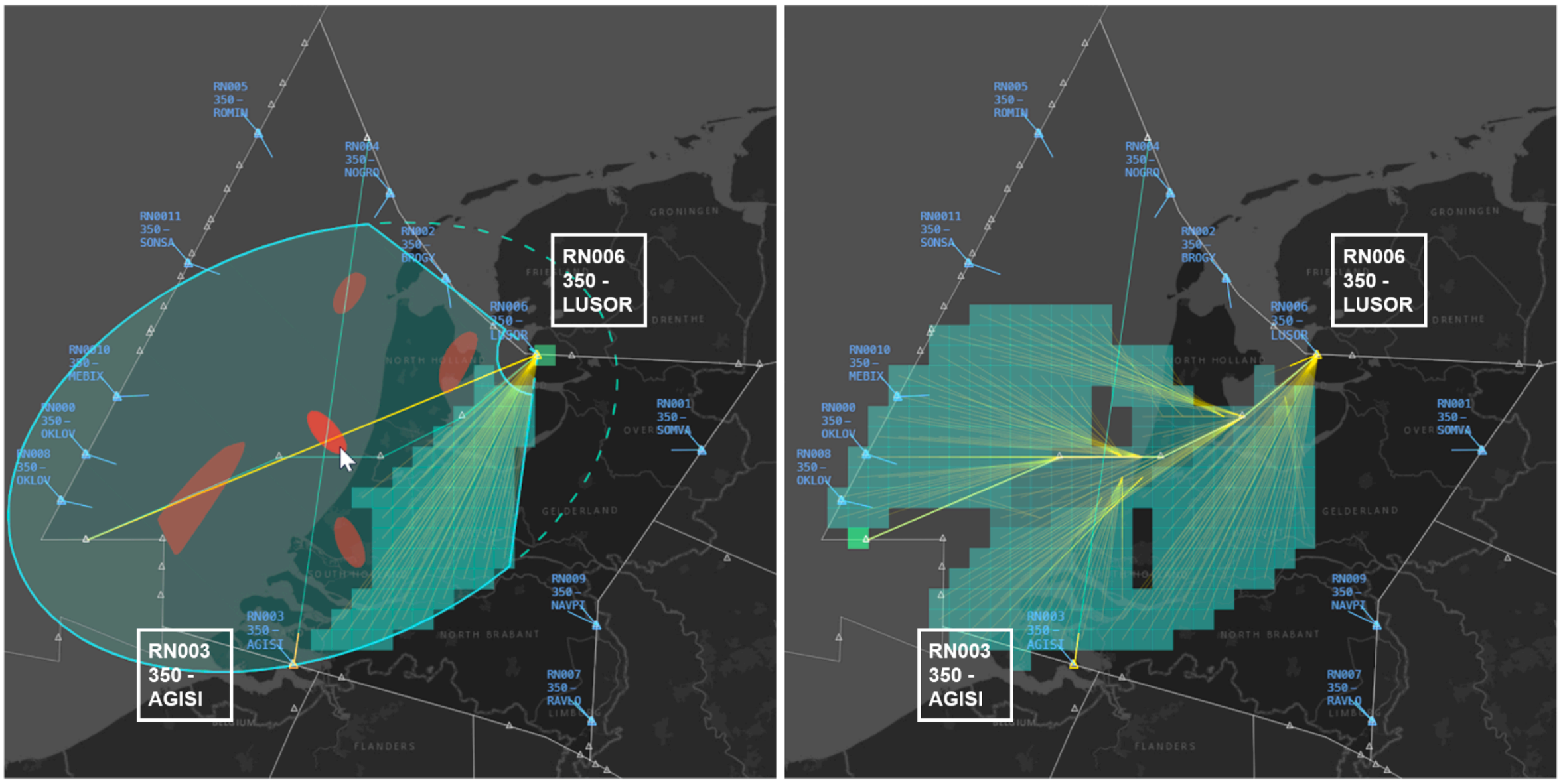

(a) First search step. (b) Final search step.

**Fig. 11.** Screenshots of the internal process of SSPP using conflict zones.

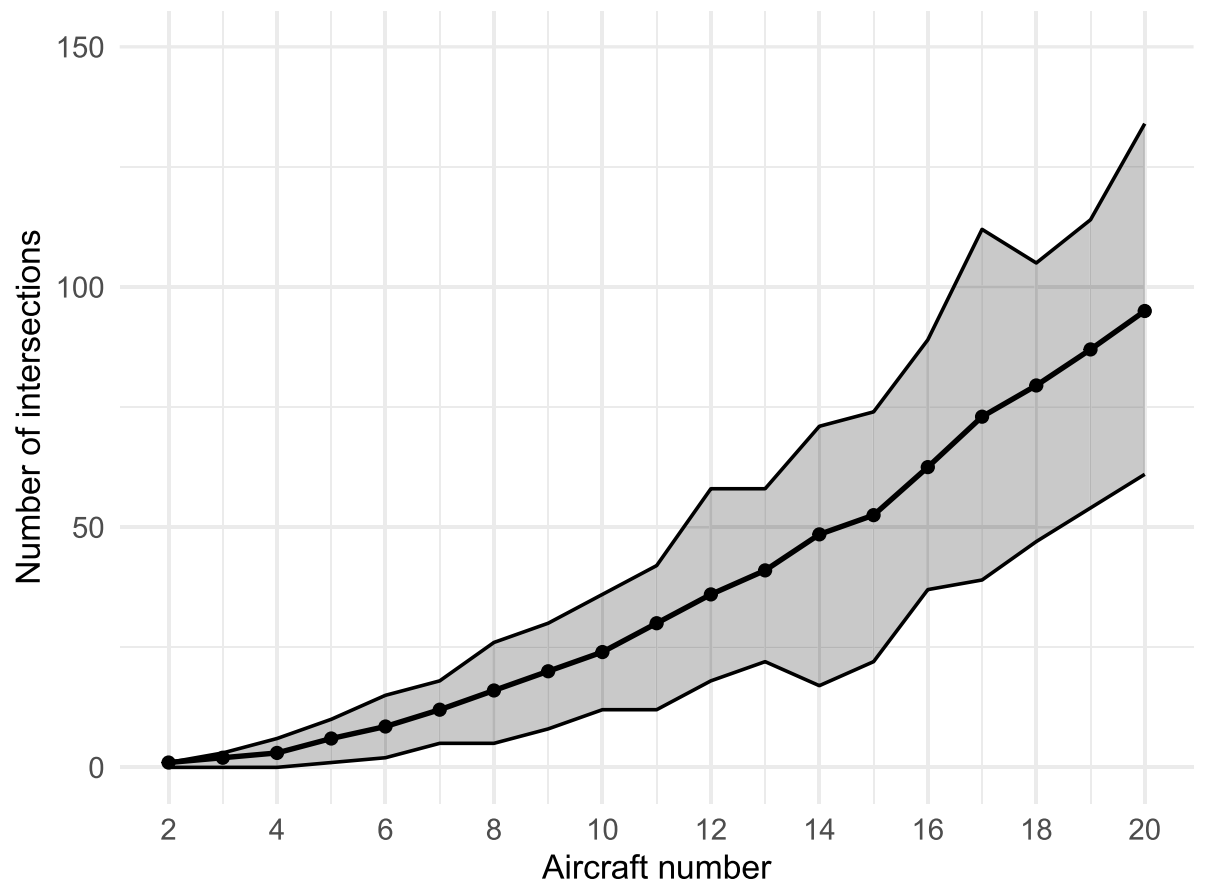

**Fig. 12.** Minimum, median, and maximum number of intersections prior to rerouting. The shaded region indicates the range between minimum and maximum values.

Unless otherwise specified, each configuration is evaluated using 20 aircraft, all operating at the same flight level. For each configuration, 50 random scenarios are generated. The number of aircraft is set to 20 because, after applying the first filtering rule for entry and exit points, only 27 usable points remain in the Delta sector for scenario generation. Moreover, since all aircraft operate at the same flight level (FL350) and their start-target pairs are randomly assigned, the resulting flight trajectories can be highly complex with many intersections, representing extreme and challenging traffic conditions. As presented in Fig. 12, before rerouting, when the number of aircraft reaches 20, the number of intersections can increase to around 100. Scenarios with larger numbers of aircraft are not constructed by extending smaller ones; instead, each is created independently from scratch.

### *6.2.1. Number of aircraft*

Fig. 13 presents the results as the number of aircraft increases from 2 to 20. In general, higher traffic density leads to greater delay, as more aircraft need to deviate from their direct routes. With 20 aircraft, SSPPV-D and SSPPE-D achieve success rates of 99.60% and 99.65%, with mean delays of 12.57 and 11.99 s, respectively. The zone-based variants achieve similar success rates of 98.95% and 99.20%, although with slightly higher mean delays of 15.23 and 14.84 s. The time-interval-based variants generally exhibit greater mean delays and lower success rates, with the differences becoming more pronounced as the number of aircraft increases. The higher delays and lower success rates of the Z and T variants are consistent with their more conservative conflict representations, which may exclude some paths that satisfy the continuous-time separation requirement.

The difference between SSPPV and SSPPE is more evident in runtime than in solution quality. With 20 aircraft, the mean runtimes of SSPPV-D, SSPPV-T, and SSPPV-Z are 7.29, 12.32, and 3.20 ms, respectively, compared with 216.34, 403.48, and 91.07 ms for the corresponding SSPPE variants. Retaining directed-edge states therefore incurs substantial computational cost for only a small improvement in success rate under the default settings. The zone-based conflict detection method is the fastest of the three for both search variants in this experiment.

Table 2 summarizes the runtime variability with 20 aircraft. The SSPPV variants show relatively low variability, with maximum sample-level mean runtimes of 17.11, 24.51, and 7.98 ms for the distance-based, time-interval-based, and zone-based methods, respectively. All of these values are well below 100 ms, indicating that even in the worst observed case, the SSPPV variants can still satisfy our 100 ms target for real-time interaction. In contrast, the SSPPE variants show substantially greater variability and longer runtimes, particularly SSPPE-T, making them less suitable for time-critical applications. However, the higher success rates of the SSPPE variants suggest that these methods may serve as fallback options when the SSPPV variants fail to identify a feasible solution.

### *6.2.2. Grid size*

Fig. 14 shows the trade-off between spatial resolution and computational cost. Increasing the grid size reduces the search space and substantially accelerates all six algorithms. For example, when the grid size changes from 2 to 5 nmi, the mean runtime of SSPPV-Z decreases from 6.30 to 0.48 ms and that of SSPPE-Z from 296.32 to 3.41 ms. However, their success rates drop from 99.15% to 96.15% and from 99.35% to 96.80%, respectively, and the mean delay increases. Coarser

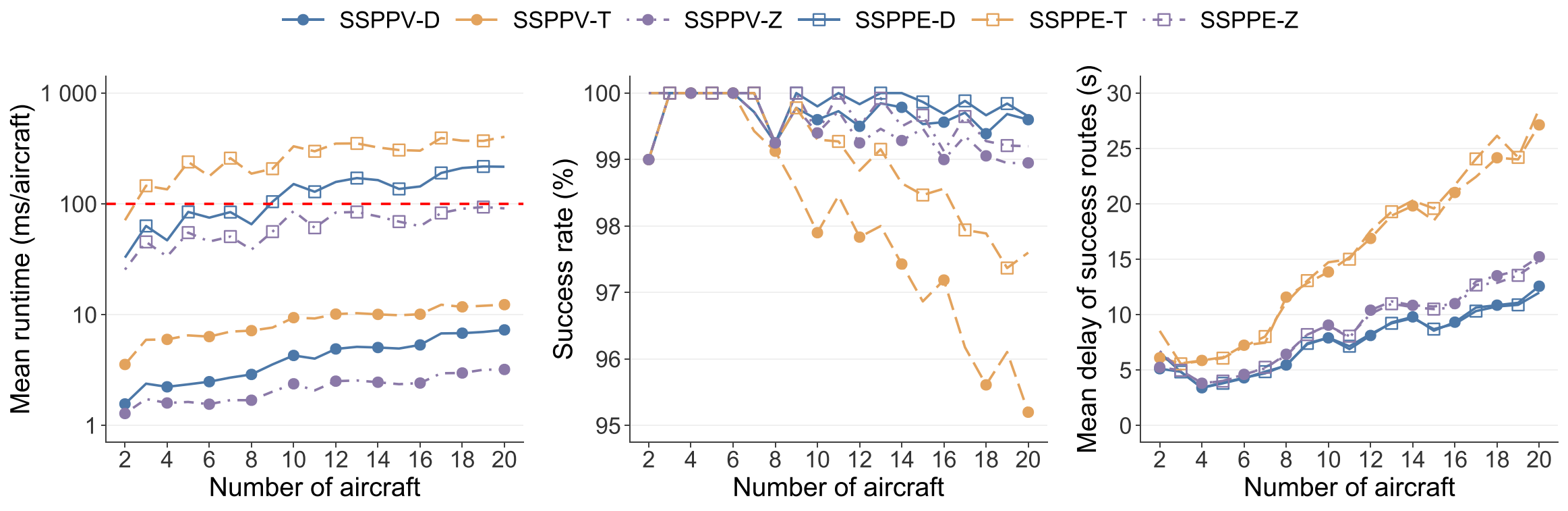

**Fig. 13.** Algorithm performance under different numbers of aircraft.

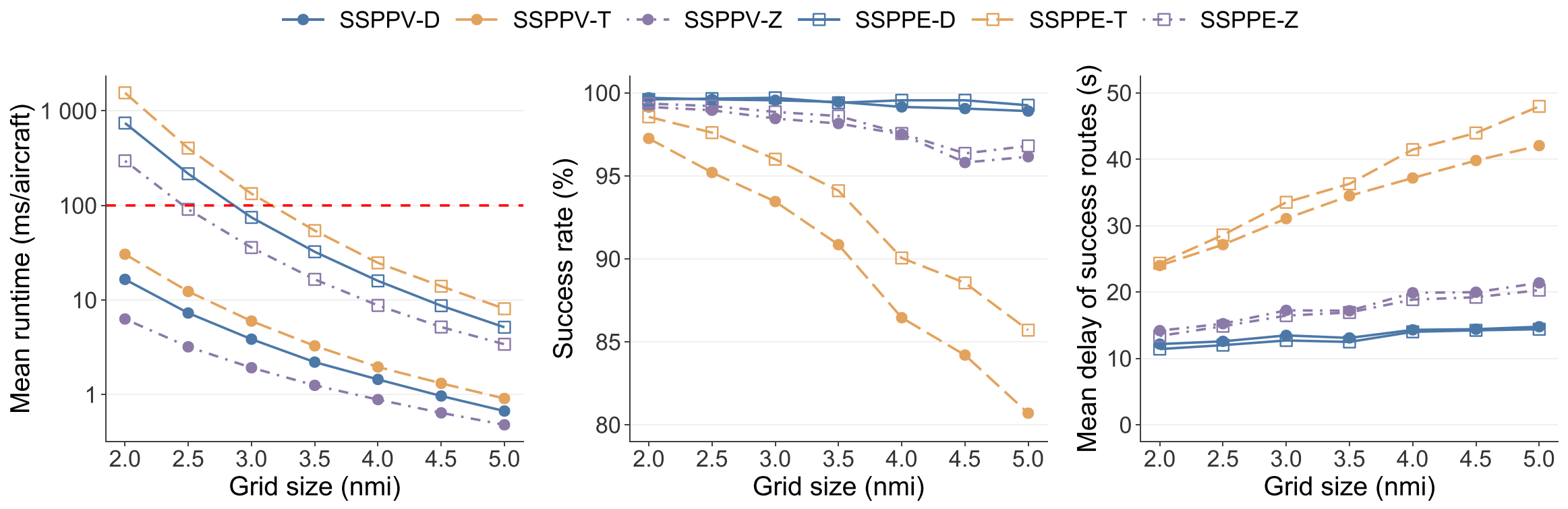

**Fig. 14.** Algorithm performance under different grid sizes.

**Table 2**
Runtime statistics (ms/aircraft) under the default settings. The reported statistics are calculated across 100 scenario-condition samples, where each sample value is the mean runtime over 20 aircraft.

| Algorithm | Mean | SD | 95% CI | Maximum |
|---|---|---|---|---|
| SSPPV-D | 7.29 | 3.46 | [6.60, 7.98] | 17.11 |
| SSPPV-T | 12.32 | 4.88 | [11.35, 13.28] | 24.51 |
| SSPPV-Z | 3.20 | 1.70 | [2.87, 3.54] | 7.98 |
| SSPPE-D | 216.34 | 164.52 | [183.69, 248.98] | 640.74 |
| SSPPE-T | 403.48 | 281.12 | [347.70, 459.26] | 1667.68 |
| SSPPE-Z | 91.07 | 75.12 | [76.16, 105.97] | 309.48 |

grids provide fewer alternatives for navigating around conflicts and obstacles.

The sensitivity to grid size varies across the conflict detection methods used in path planning. The distance-based variants maintain success rates above 98% across all tested grid sizes. In contrast, the time-interval-based variants are more sensitive: as the grid size increases from 2 to 5 nmi, the success rate drops from 97.25% to 80.70% for SSPPV-T and from 98.55% to 85.70% for SSPPE-T. The larger cells cover wider areas, making the time-interval-based method more conservative and reducing the number of candidate paths. Thus, the runtime reduction achieved with larger grid sizes should be considered together with the corresponding loss in routing success.

### 6.2.3. *Turning angle limit*

Fig. 15 illustrates how algorithm performance changes as the turning angle limit is gradually relaxed. A larger turning angle limit allows the algorithm to consider a wider range of aircraft maneuvers and generally leads to higher success rates. For example, increasing the limit from 30° to 60° raises the success rate of SSPPV-D from 93.95% to 99.60% and that of SSPPV-Z from 92.70% to 98.95%. When the angle is restricted to 30°, SSPPE-D and SSPPE-Z achieve higher success rates of 97.15% and 95.55%, respectively, than their corresponding SSPPV variants. This is because SSPPE retains different incoming headings at the same location through its directed-edge state representation, allowing more possible subsequent maneuvers under stricter turning angle constraints.

A larger turning angle limit also increases runtime because more candidate maneuvers can be considered at each expansion. When the turning angle limit is below 60°, SSPPE-Z can find paths within 100 ms for real-time interaction. Beyond 60°, the distance-based and zone-based variants show only marginal improvements in success rate, whereas the time-interval-based variants continue to benefit from the additional maneuvering freedom. Therefore, the default limit of 60° provides a reasonable balance between algorithm performance and operational realism. The lower mean delay observed under a stricter limit does not necessarily indicate better path quality, since more difficult cases that fail to find a conflict-free path are excluded from the mean delay calculation.

### 6.2.4. *Number of rerouting points*

Fig. 16 compares the distributions of runtime and success rate with a maximum of 1 or 5 rerouting points. Each box contains 100 samples, obtained from 50 traffic scenarios with and without a no-fly zone. With only one rerouting point, SSPPV and SSPPE achieve the same success rates and delays under each conflict detection method. Under this setting, every intermediate point is reached directly from the start point, making the state representations of SSPPV and SSPPE equivalent for path planning.

Allowing more rerouting points substantially improves the success rates. For SSPPV-D and SSPPV-Z, the success rates increase from 94.15% to 99.60% and from 91.25% to 98.95%, respectively, while their mean runtimes increase only slightly. SSPPE also benefits from the additional rerouting points, but at a much higher computational cost. For example, the success rate of SSPPE-Z increases from 91.25% to 99.20%, while its mean runtime increases from 7.34 to 91.07 ms. These results demonstrate the benefit of allowing multiple rerouting points in complex traffic, while also highlighting the computational cost of using directed-edge states in deeper searches.

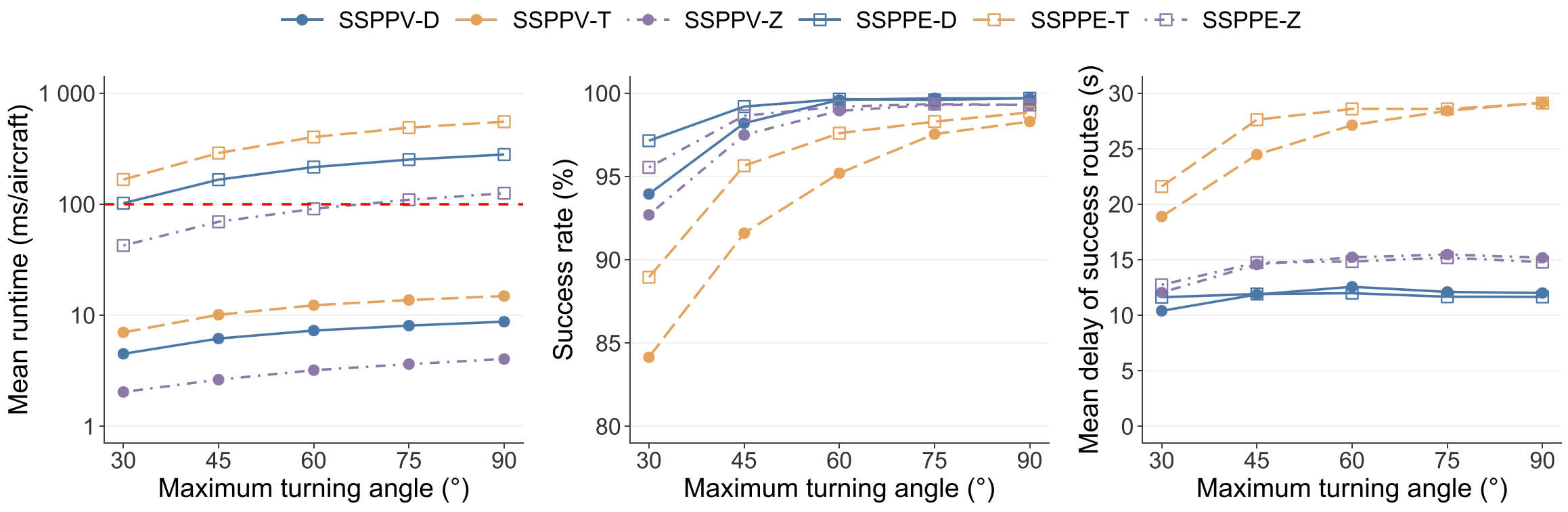

**Fig. 15.** Algorithm performance under different turning angle limits.

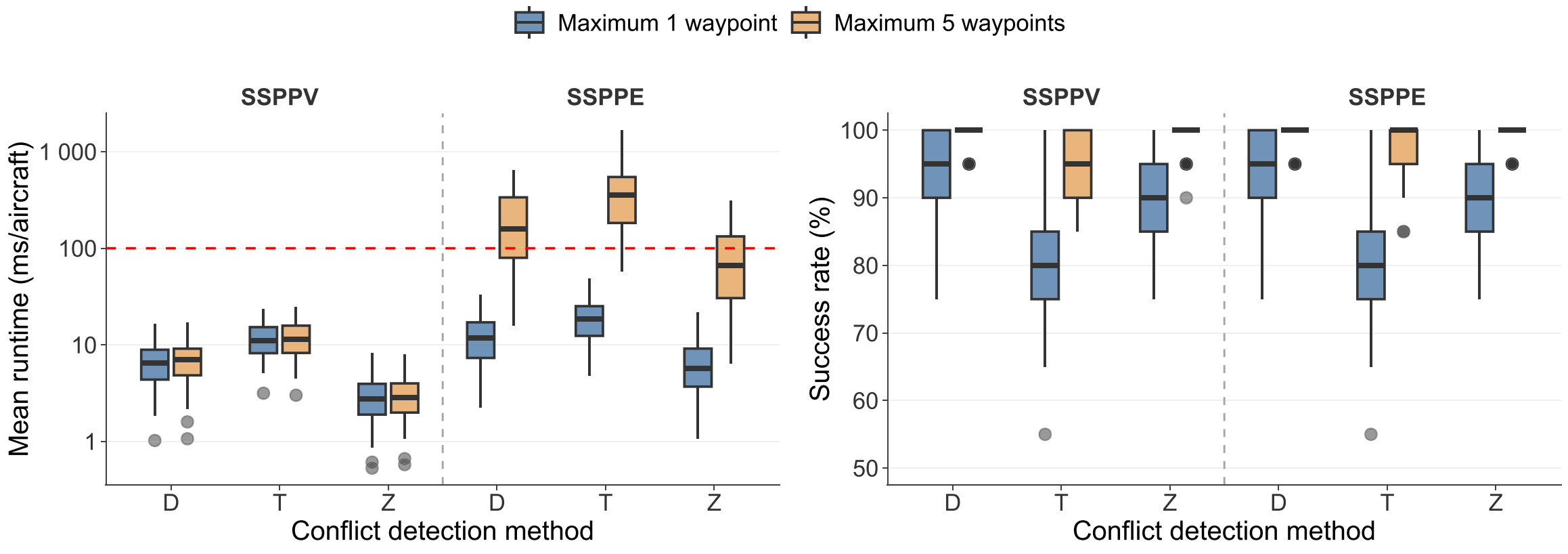

**Fig. 16.** Algorithm performance with a maximum of 1 or 5 rerouting points.

#### *6.2.5. Planning order*

As aircraft are planned sequentially, earlier decisions determine the dynamic obstacles encountered by later aircraft. To evaluate this order dependence, each of the 50 scenarios is tested using 50 random planning orders, both with and without a no-fly zone, resulting in 100 samples and 5000 planning-order evaluations per algorithm. Fig. 17 summarizes the variation across planning orders for each sample. Runtime sensitivity is measured by the ratio of the 90th to the 10th percentile, while success-rate sensitivity is measured by the difference between the maximum and minimum success rates. Each box therefore contains 100 sensitivity values, one derived from each sample.

SSPPV is less sensitive to planning order in terms of runtime. The median P90/P10 ratios range from 1.47 to 1.57 for SSPPV, compared with 2.26 to 2.53 for SSPPE. In some scenarios, the ratios for SSPPE-T and SSPPE-Z exceed 10, indicating that an unfavorable planning order can considerably increase the search effort for SSPPE. The distance-based variants have a median success-rate range of zero, whereas the time-interval-based variants have median ranges of 15 and 10 percentage points, respectively. Therefore, planning order affects both runtime and routing success, with the time-interval-based method being the most sensitive due to its conservative conflict representation.

#### *6.2.6. Shadowcasting vs. line-of-sight checking*

Fig. 18 compares shadowcasting (SC) with line-of-sight checking (LOS), while keeping the search framework (SSPPV or SSPPE) and conflict detection method (D, T, or Z) unchanged. Shadowcasting reduces runtime for all six variants, with average speedups over LOS of 3.71 for SSPPV and 3.38 for SSPPE across the three conflict detection methods. For example, SSPPV-Z requires 3.20 ms with SC and 15.87 ms with LOS, while SSPPE-Z requires 91.07 and 406.12 ms, respectively.

The advantage arises from computing visibility over a region in a single scan. This is particularly beneficial when many candidate paths share grid cells near the current node. The zone-based method can also incorporate the rasterized conflict front into the same scan, further reducing the need for separate conflict checks. These results support the use of shadowcasting for generating a full solution space rather than evaluating each candidate path independently.

#### *6.2.7. Solution space vs. local neighborhoods*

Since SSPP follows the A* search framework, the key difference lies in node expansion: traditional A* considers eight neighboring nodes at each step, whereas SSPP considers the full solution space. Therefore, we compare SSPP with conflict-free A* using square neighborhoods of radius $k$ cells, as shown in Fig. 19. Shadowcasting, conflict detection methods, and all routing constraints remain unchanged. The neighborhood radius $k$ starts at 4 rather than 1, since $4 \times 2.5 = 10\,\text{nmi}$, corresponding to the minimum waypoint distance constraint. Thus, the comparison isolates the effect of neighborhood size from other algorithmic factors.

The results indicate that a small neighborhood substantially reduces routing success. At $k = 4$, success rates are only 9.10%–11.30% across the six variants. As the neighborhood size increases, routing success improves markedly. At $k = 12$, E-D achieves a success rate of 99.10% with a mean runtime of 109.53 ms, compared with 99.65% and 216.34 ms for the full solution space (SS). Runtime, in contrast, does not vary monotonically with neighborhood size. Although a smaller candidate set reduces the computational cost of each expansion, it may require more intermediate states or lead to unsuccessful searches. The low mean delay for small neighborhoods should be interpreted with their low success rates. The results also highlight the advantages of the zone-based method. For the zone-based variants, using the full solution space achieves a runtime comparable to that with $k = 12$, while providing higher success rates and lower mean delays. Overall, an appropriate neighborhood size can balance runtime and routing success, whereas the full solution space retains a wider range of maneuvering alternatives.

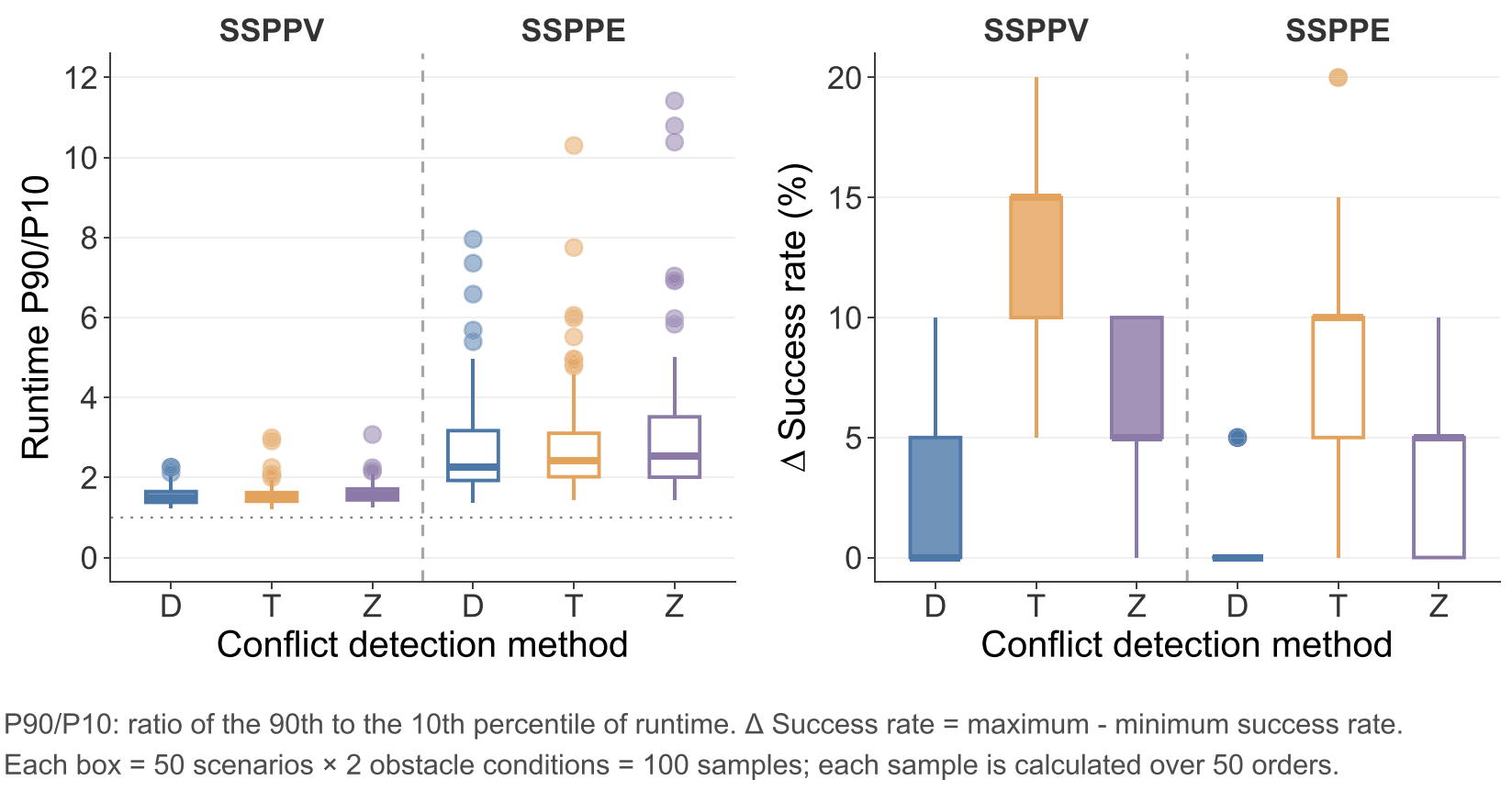

**Fig. 17.** Algorithm performance under different planning orders.

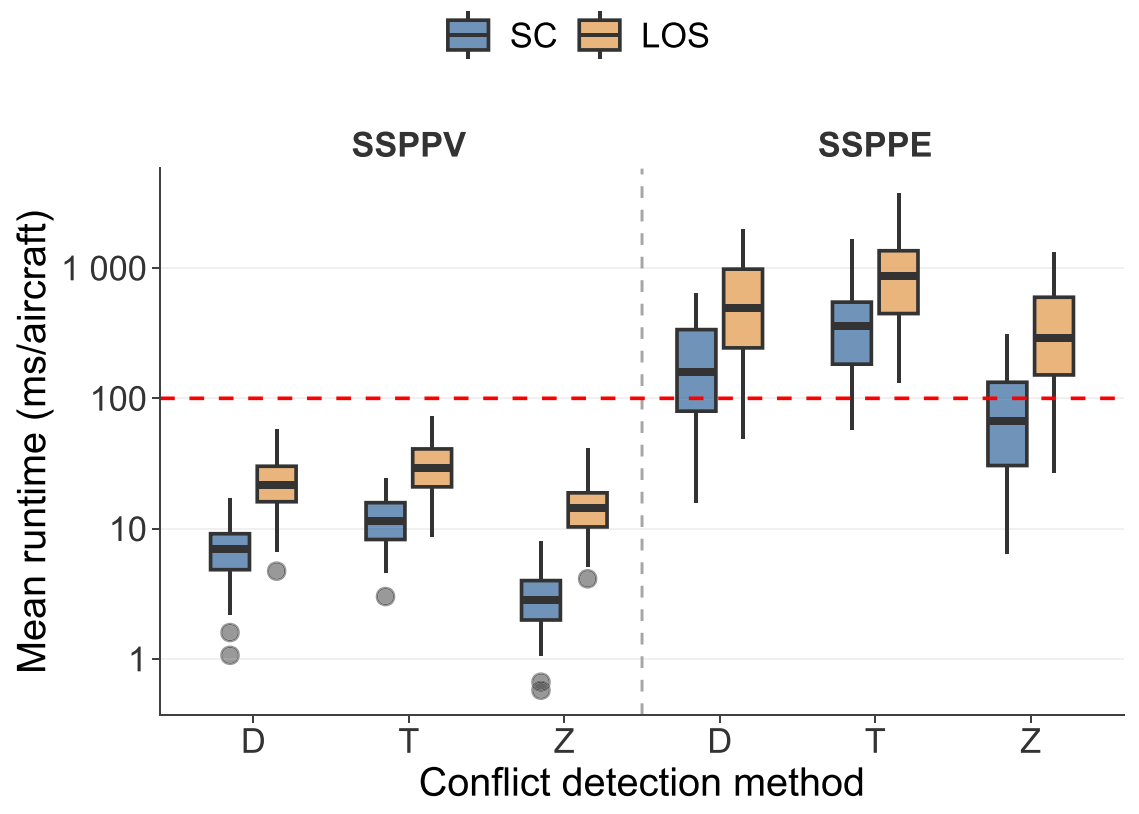

**Fig. 18.** Comparison of shadowcasting and line-of-sight checking for SSPP.

*6.2.8. Trade-off between delay and separation*

Fig. 20 presents the trade-off between delay and separation for SSPPV-D and SSPPE-D in scenarios with 2 and 20 aircraft. The separation weight is $w_2 = 1 - w_1$ in Eq. (12). A weight of zero prioritizes delay, whereas a weight of one prioritizes separation. The separation metric is the mean of the pairwise minimum separations over all aircraft pairs, including the direct fallback routes of aircraft for which rerouting is unsuccessful.

With two aircraft, increasing the separation weight produces a clear increase in separation at the cost of greater delay. For SSPPV-D, increasing the weight from 0 to 0.8 raises the mean separation from 36.34 to 43.12 nmi, while the mean delay of successful routes increases from 5.12 to 22.73 s. At a weight of one, these values reach 55.25 nmi and 101.98 s, respectively. SSPPE-D shows a similar trend, with slightly greater separation but also greater delay at the highest weight.

With 20 aircraft, the increase in mean separation is smaller relative to the additional delay. For SSPPV-D, increasing the weight from 0 to 0.8 raises the mean separation from 39.20 to 41.11 nmi, while the mean delay of successful routes increases from 12.57 to 74.02 s. This is because earlier aircraft have greater flexibility to increase separation, whereas later aircraft must accommodate previously assigned trajectories. A weight of one results in particularly large delays for both algorithm variants, highlighting the importance of retaining a delay penalty.

The weighted objective also increases computational cost. With 20 aircraft, increasing the separation weight from 0 to 0.8 raises the mean runtime of SSPPV-D from 7.29 to 31.91 ms per aircraft and that of SSPPE-D from 216.34 to 1,483.47 ms. At a weight of 0.8, both variants achieve similar mean separations of approximately 41.11 nmi, despite the substantial difference in runtime. Overall, intermediate weights provide a more balanced trade-off between delay and separation, while SSPPV offers lower computational cost for interactive adjustment of the objective.

## 7. Discussion

*7.1. Comparison of conflict detection methods and algorithm variants*

Under the default settings with 20 aircraft and a 2.5 nmi grid, the three conflict detection methods exhibit distinct performance characteristics. Zone-based detection achieves the lowest runtime, with mean speedups of 2.48 and 5.37 over the distance-based and time-interval-based methods, respectively, when averaged across SSPPV and SSPPE. Distance-based detection generally provides higher routing success and lower delay. Time-interval-based detection is comparatively more conservative, resulting in higher runtime and lower success rates.

These differences suggest that the overall algorithm performance is determined not only by the computational cost of conflict checks, but also by how each method represents conflicts and consequently constrains the feasible search space. This is particularly evident for the time-interval-based method, whose routing success degrades more noticeably with larger grid cells because the occupancy representation becomes increasingly conservative. In contrast, the zone-based method converts dynamic conflicts into a form that can be directly incorporated into shadowcasting scans, substantially reducing computational cost while maintaining routing performance close to distance-based detection.

The benefit of retaining directed-edge states depends on the path-planning constraints. Tighter turning-angle limits make the ability of SSPPE to retain different incoming headings more beneficial, lead-ing to higher routing success than SSPPV. However, this comes at a substantial computational cost. Under the default settings with 20 aircraft and a 2.5 nmi grid, SSPPV is on average 27.49 times faster than SSPPE across the three conflict detection methods. SSPPV therefore offers substantially lower computational cost with comparable planning performance, whereas SSPPE may be advantageous when higher routing success is preferred despite the additional computational cost.

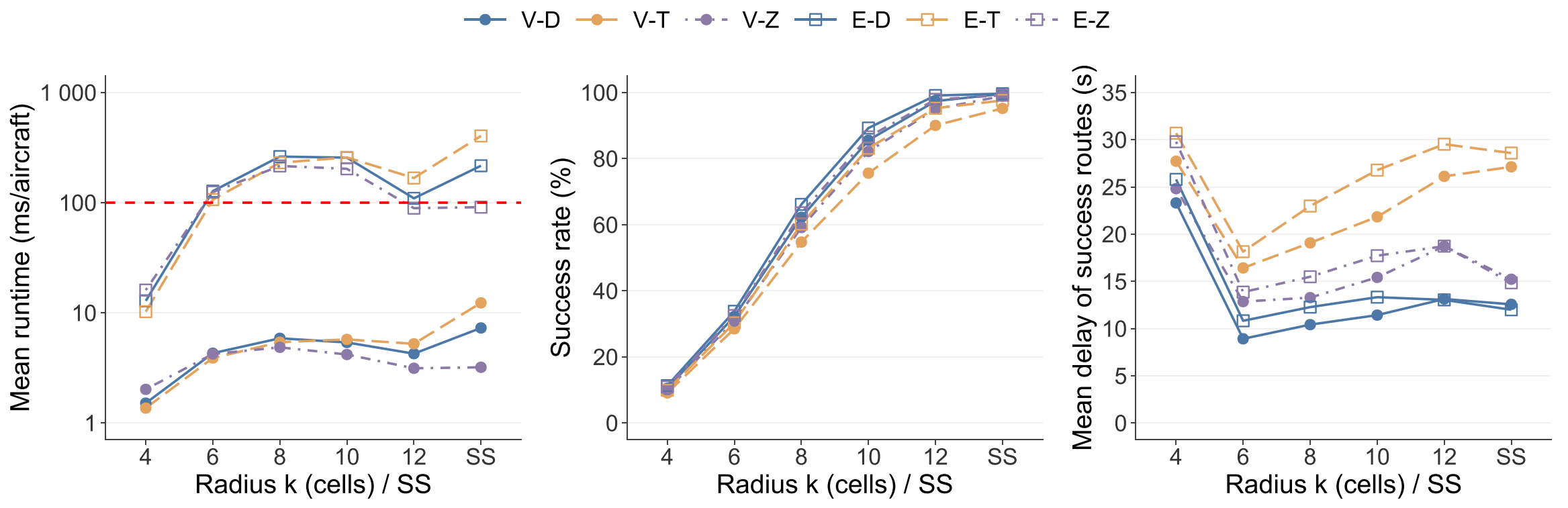

**Fig. 19.** Comparison of SSPP with solution space and conflict-free A* with square neighborhoods of radius $k$ cells. V and E denote vertex-based and edge-based A* search, respectively, while D, T, and Z denote the three conflict detection methods. SS represents the full solution space.

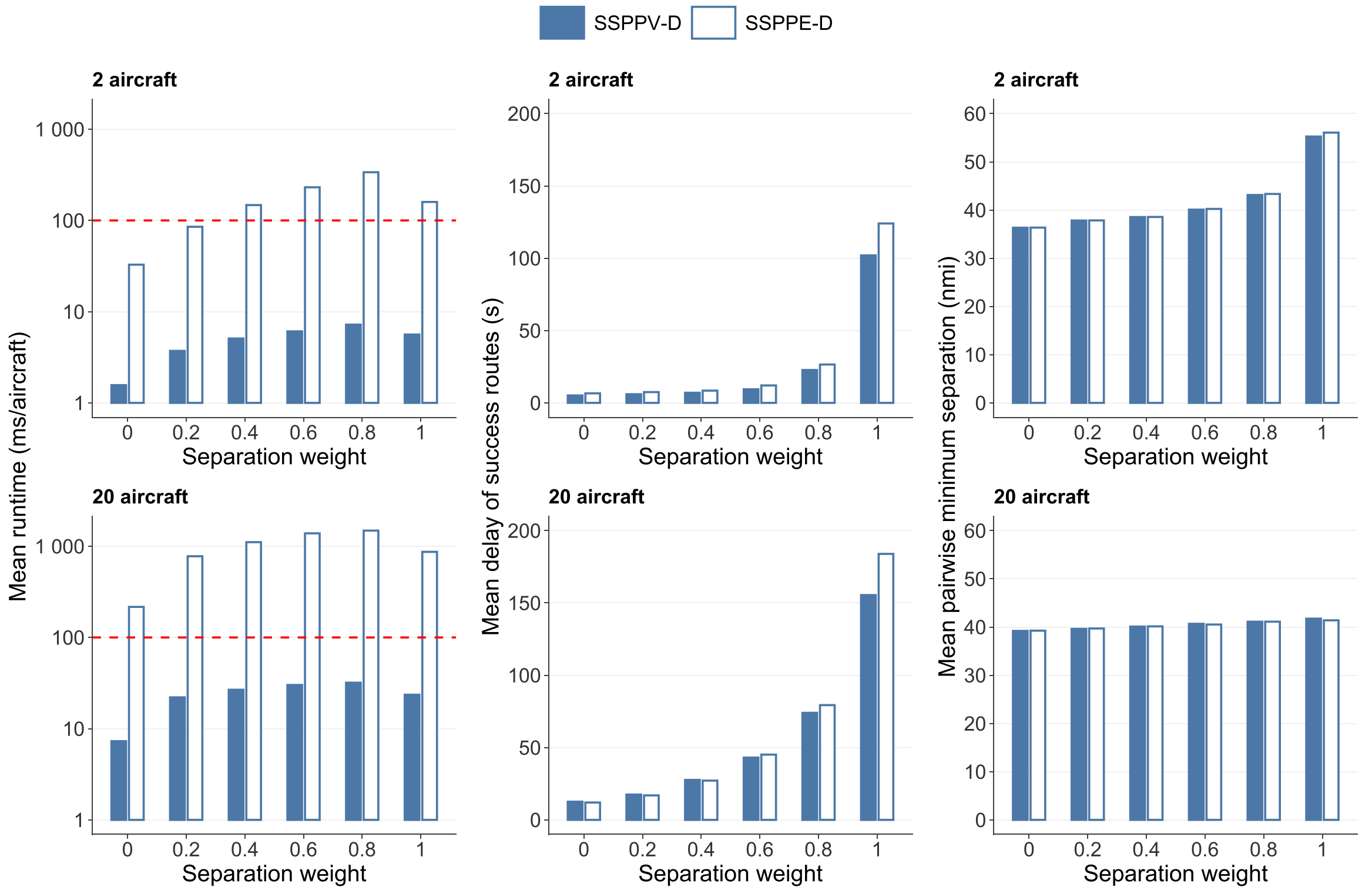

**Fig. 20.** Algorithm performance under different weights for the trade-off between delay and separation. Mean pairwise minimum separation is defined as the average of the minimum separation distances between all aircraft pairs.

The weighted variant demonstrates a clear trade-off between delay and separation. The improvement in separation becomes smaller in denser traffic, while a delay penalty remains important to avoid excessive delay. Moreover, the additional computational cost of edge-based search does not yield a comparable improvement in separation, suggesting that retaining multiple incoming-heading alternatives provides limited benefit for this objective in the tested scenarios. SSPPV-D therefore provides a more practical choice for the weighted formulation.

### *7.2. Limitations and future research*

When designing algorithms for SSPP, we have three key assumptions. First, we focus on 2D rather than 3D path planning because the airspace is divided into flight levels, which decouples the altitude dimension from the horizontal dimension. Second, we assume a constant aircraft speed for path planning, as heading changes are generally preferred over speed changes for conflict resolution in en-route ATC. Third, we ignore aircraft turning dynamics and trajectory uncertainties, which is a common assumption used to simplify the path-planning problem.

To extend the applicability of SSPP to more complex environments, these assumptions can be readily relaxed or compensated for. Relaxing the first two assumptions naturally introduces two additional conflict resolution mechanisms: altitude adjustments and speed adjustments. Therefore, we can define an ordered sequence of conflict resolution strategies: for example, first attempt heading changes; if that fails, ad-just flight levels; and if conflicts still remain, modify speed. In this way, a three-layer solution space can be constructed. The first layer corresponds to the solution space considered in this article, where conflicts are resolved using only heading changes. The second layer focuses on cells that are unreachable in the first layer and attempts to make them reachable through altitude adjustments. The third layer addresses cells that remain unreachable after altitude adjustments by modifying speed. The remaining search procedures for path planning are the same. The approach described here is only one potential extension. Future research could examine in depth how to extend the solution space concept and solution space path planning to 3D, as well as how to incorporate speed adjustments.

The third assumption can be addressed by adding a buffer to the minimum separation standard. While this method may introduce additional delay, it is the simplest way to mitigate the issue. In practice, ATCos also deal with uncertainties by incorporating separation buffers in their decisions. When ATCos use SSPP as a decision-support tool, they should be aware of this assumption and continuously monitor aircraft trajectories to ensure safety. They may also be allowed to adjust the separation buffer as needed.

Future work could evaluate SSPP under more realistic scenarios by incorporating additional environmental and system-related factors. For example, space weather may degrade communication, navigation, and surveillance services through disturbances in the ionosphere and near-Earth space environment, while enhanced radiation may affect onboard systems and avionics and increase radiation exposure for crew and passengers [74,75]. When these effects lead to operational restrictions on particular routes, regions, altitudes, or time periods, they could be represented in SSPP as additional spatial or spatiotemporal constraints.

To further validate the effectiveness of SSPP and facilitate its real-world application, future research could involve human-in-the-loop experiments to explore how ATCos collaborate and interact with SSPP in operations, and whether its intrinsically interpretable tree-based structure, rendered solution space representation, and low computational latency will benefit operator understanding, usability, acceptance, and trust. This involves not only algorithm design, but also the design of human-machine interaction and interfaces. The case study results highlight the importance of the order in conflict-free path planning (conflict resolution), which is also an interesting direction for future research.

## 8. Conclusion

This paper presents Solution Space Path Planning (SSPP), including multiple algorithmic variants, for supporting en-route Air Traffic Control (ATC). SSPP was developed from a human-centered perspective, in which feasible actions and their operational constraints are explicitly represented through interpretable solution spaces, while computational efficiency is maintained for real-time interaction and supervision. The algorithm combines the advantages of solution space representations and any-angle path planning, offering an intrinsically interpretable algorithmic structure with straight and low-delay paths for aircraft routing. To accelerate visibility checks with static obstacles, shadowcasting is adapted and integrated into the solution space generation process. For addressing dynamic obstacles, three conflict detection methods have been introduced: distance-based, time-interval-based, and zone-based.

Under the default settings with 20 aircraft and a 2.5 nmi grid, the distance-based method achieves success rates of 99.60% for SSPPV-D and 99.65% for SSPPE-D, with mean runtimes of 7.29 and 216.34 ms per aircraft, respectively. The time-interval-based method has the lowest success rates, at 95.20% for SSPPV-T and 97.60% for SSPPE-T, as well as the highest runtimes. The zone-based method maintains relatively high success rates of 98.95% for SSPPV-Z and 99.20% for SSPPE-Z, while being on average 2.48 and 5.37 times faster than the distance-based and time-interval-based methods, respectively, across SSPPV and SSPPE.

Building on the solution space framework and A* search, two SSPP variants, SSPPV and SSPPE, are developed. Under the default settings with 20 aircraft, SSPPV achieves similar solution quality to SSPPE but offers about 27.49 times faster computation. From a human-centered design perspective, such low computational latency is particularly desirable because it can support responsive and uninterrupted interaction, including iterative exploration and what-if probing. Considering the results across all evaluated dimensions, the zone-based SSPPV variant (SSPPV-Z) emerges as the most promising configuration, combining low runtime, a high success rate, low aircraft delay and an intrinsically interpretable conflict-zone representation.

An extension of SSPP has also been developed using distance-based conflict detection, focusing on the trade-off between delay and separation, to demonstrate its potential for improving solution robustness. Given the interpretable structure of solution spaces, the deterministic nature of search algorithms, and the ability to generate operationally compliant routes, SSPP provides a promising basis for human-centered aviation automation. Future research should investigate through human-in-the-loop experiments how these design properties affect operator understanding, usability, acceptance, and trust.

## CRediT authorship contribution statement

**Yiyuan Zou:** Writing – review & editing, Writing – original draft, Visualization, Validation, Software, Methodology, Investigation, Funding acquisition, Formal analysis, Data curation, Conceptualization; **Wenying Lyu:** Writing – original draft, Visualization, Validation, Investigation, Data curation, Conceptualization; **Clark Borst:** Writing – review & editing, Validation, Supervision, Resources, Project administration, Investigation, Funding acquisition, Conceptualization.

## Declaration of competing interest

The authors declare that they have no known competing financial interests or personal relationships that could have appeared to influence the work reported in this paper.

## Acknowledgements

The authors would like to thank the China Scholarship Council for its financial support (No. 202106830036).

## Data availability

Data will be made available on request.